\documentclass[11pt]{article}

\usepackage[margin=1in]{geometry}
\usepackage[utf8]{inputenc} 
\usepackage[T1]{fontenc}    
\usepackage{hyperref}       
\usepackage{url}            
\usepackage{booktabs}       
\usepackage{amsfonts}       
\usepackage{nicefrac}       
\usepackage{microtype}      
\usepackage{xcolor}         

\usepackage{amsmath,mathrsfs}
\usepackage{dsfont}
\usepackage{amssymb}
\usepackage{amsthm}
\usepackage{bm}
\usepackage{verbatim}
\usepackage{mathtools}
\mathtoolsset{showonlyrefs}
\usepackage{multirow}
\usepackage{subfigure}
\usepackage{float}
\usepackage{tikz}
\usepackage{pgfplots}
\usepackage{algorithm}
\usepackage{algorithmicx}
\usepackage{algpseudocode}

\newcommand{\field}[1]{\ensuremath{\mathbb{#1}}}
\newcommand{\R}{\ensuremath{\field{R}}} 
\newcommand{\1}{\ensuremath{\mathbf{1}}} 
\newcommand{\I}[1]{\ensuremath{\mathbb{I}_{\left\{#1\right\}}}} 
\newcommand{\PR}{\ensuremath{\mathsf{P}}} 
\newcommand{\E}{\ensuremath{\mathsf{E}}} 
\newcommand{\defeq}{\ensuremath{\triangleq}}

\newcommand{\Fscr}{\ensuremath{\mathcal F}}

\newcommand{\Xscr}{\ensuremath{\mathcal X}}

\DeclareMathOperator{\Var}{Var}

\newtheoremstyle{thm-sf}{}{}{\itshape}{}{\sffamily\bfseries}{.}{ }{}
\theoremstyle{thm-sf}
\newtheorem{theorem}{Theorem}
\newtheorem{proposition}{Proposition}

\newtheorem{lemma}{Lemma}
\theoremstyle{definition}

\theoremstyle{remark}

\title{Debiasing Samples from Online Learning Using Bootstrap}
\author{Ningyuan Chen,\thanks{The Rotman School of Management, University of Toronto; ningyuan.chen@utoronto.ca} \quad 
Xuefeng Gao\thanks{Department of Systems Engineering and Engineering Management, The Chinese University of Hong Kong; xfgao@se.cuhk.edu.hk} \quad 
Yi Xiong,\thanks{Department of Systems Engineering and Engineering Management, The Chinese University of Hong Kong; yxiong@se.cuhk.edu.hk}}

\begin{document}

\maketitle

\begin{abstract}
	It has been recently shown in the literature \cite{nie2018adaptively,shin2019sample,shin2019bias} that the sample averages from online learning experiments are biased when used to estimate the mean reward. To correct the bias, off-policy evaluation methods, including importance sampling and doubly robust estimators,  typically calculate the conditional propensity score, which is ill-defined for non-randomized policies such as UCB. This paper provides a procedure to debias the samples using bootstrap, which doesn't require the knowledge of the reward distribution and can be applied to any adaptive policies.  Numerical experiments demonstrate the effective bias reduction for samples generated by popular multi-armed bandit algorithms such as Explore-Then-Commit (ETC), UCB, Thompson sampling (TS) and $\epsilon$-greedy (EG). We analyze and provide theoretical justifications for the procedure under the ETC algorithm, including the asymptotic convergence of the bias decay rate in the real and bootstrap worlds.
\end{abstract}

\section{Introduction}\label{sec:introduction}

Online learning and specifically the multi-armed bandit problem, has seen great success in many applications, including recommender systems \cite{li2011unbiased}, clinical trials \cite{villar2015multi}, dynamic pricing \cite{den2015dynamic}, and A/B testing or online experiments \cite{burtini2015survey}.
The primary goal of online learning is to explore available options (arms) while maximizing the payoff (or equivalently minimize the regret) at the same time.
Many algorithms or policies achieve the goal remarkably well, including UCB (Upper Confidence Bound) \cite{lai1985asymptotically} and Thompson sampling \cite{thompson1933likelihood}.
After the conclusion of the experiment and the collection of the data, typically additional evaluations need to be conducted based on the collected data without running new experiments.
This is usually referred to as ``off-policy evaluation'' \cite{precup2000eligibility,li2011unbiased,li2012unbiased,jiang2016doubly}. 

\textbf{Motivating Example.} To illustrate the motivation, consider the following hypothetical example in early-stage clinical trials.
The online learning framework has been applied to Phase I/II trials to identify the optimal dose \cite{villar2015multi,chen2020adaptive,aziz2021multi}.
Suppose a clinical trial is conducted on 100 patients sequentially to evaluate the efficacy of two drugs, $A$ and $B$.
This can be treated as a multi-armed bandit problem with two arms.
The manager applies the UCB algorithm to allocate the drugs to the patients while observing their responses in terms of health measures.
Eventually, the responses of the 100 patients are recorded, among which 90 patients are allocated drug $A$ and the remaining 10 patients are allocated drug $B$.
Apparently, the UCB algorithm determines that drug $A$ is more efficacious than $B$.
Suppose, in addition, the manager wants to evaluate the mean efficacy of the inferior drug $B$ without running new trials.
One natural step is to take the sample average of the responses from the 10 patients who have been allocated drug $B$.
Does the quantity reflect the actual efficacy of drug $B$?

The answer is no, which has been documented in a few recent papers \cite{nie2018adaptively,shin2019sample,shin2019bias}.
To see the intuition, note that UCB tends to choose the arm with a higher empirical average.
Therefore, if during the trial the sample average of drug $B$ is less than its actual mean efficacy, then UCB tends to choose drug $A$ more often and the realized negative bias is not likely to be fully corrected.
On the other hand, if the sample average happens to be higher than the actual mean, then UCB tends to choose drug $B$ relatively more and correct the realized positive bias.
Due to the asymmetry, the sample average of drug $B$ is typically negatively biased.

To correct the bias, the literature on off-policy evaluations provides a few potential methods \cite{li2015toward,swaminathan2015counterfactual,jiang2016doubly,dimakopoulou2017estimation,farajtabar2018more,hadad2021confidence}, including importance sampling and doubly robust estimators.
Embedded in almost all the methods, there is a key concept called the propensity score, i.e., the  probability that an arm is chosen in a period under the bandit algorithm.
To estimate the unconditional propensity scores, the distribution of the rewards of the arms is required because of the adaptive nature of most bandit algorithms.
However, the lack of prior knowledge of the rewards, not even their means, is precisely the reason to resort to online learning in the first place.
For conditional propensity scores, they are readily available for policies with internal randomization such as TS and EG \cite{hadad2021confidence}.
However, for other popular policies such as ETC and UCB which are non-randomized, the methods do not work and some adjustments have to be used \cite{dimakopoulou2017estimation}.

\textbf{Our Contribution.}
To overcome the challenge, we introduce a simulation-based algorithm to debias the sample means and provide more reliable estimators for the mean rewards of the arms.
At a high level, the algorithm can be described in the following steps:
\begin{enumerate}
	\item Based on the collected data from the experiment in the real world, we construct bootstrap reward distributions, $P_1^*$, \dots, $P_K^*$, for the $K$ arms. The bootstrap distribution could be Efron's bootstrap \cite{efron1994introduction} or multiplier bootstrap \cite{van1996weak,chernozhukov2013gaussian}.
	\item Simulate independent experiments in the ``bootstrap'' world using the same policy as in the real world, say, UCB.
	The rewards of the arms in the bootstrap world are generated using $\{P_k^*\}_{k=1}^K$.
	\item For a bootstrap experiment, the difference between the sample average of the rewards of arm $k$ and the mean reward in the ``bootstrap'' world $\E_{P_k^*}[X]$, which is the sample mean of arm $k$ in the real-world experiment, provides a realization of the bias in the ``bootstrap'' world.
	\item Average the realized bias of all the bootstrap experiments and use it to debias the data in the real world.
\end{enumerate}
The algorithm has the following features.
First, it doesn't require any knowledge of the probability model that generates the rewards.
This is the benefit of using bootstrap to construct data-dependent distributions.
Second, it is not specific to certain bandit algorithms.
As long as the bandit algorithm that generates the data in the real world can be replicated in the bootstrap world, the debiasing procedure works. 
In Section~\ref{sec:numerical}, we apply the procedure to four popular bandit algorithms: Explore-Then-Commit (ETC), UCB, Thompson sampling, and $\epsilon$-greedy.
The corrected bias is always smaller, demonstrating the effectiveness of the procedure.

We also provide theoretical justifications when applying the approach to the ETC algorithm. 
For the ETC algorithm, we theoretically characterize the biases in the real world and the bootstrap world when there are two arms and the rewards are Gaussian. 
This allows us to evaluate how well the bootstrap bias approximates the real-world bias.
In particular, we show that: 

\textbf{Informal Result}: the bias of the real and bootstrap worlds decays exponentially as the length of the exploration phase grows. The ratio of their decay rates converges to one.

When the rewards are not Gaussian, we show that the ratio of the decay rate may depend on the 
tail behavior of the reward distribution, in particular, the Legendre-Fenchel transform of the reward distribution.
We provide an asymptotic bound for the ratio.

Finally, we point out that in contrast to existing studies (e.g. \cite{shin2019bias}) which provide bounds on the bias for general bandit algorithms, we are able to give precise estimates of the bias under the ETC algorithm.

\textbf{Related Work.}
In online learning problems, reward samples are collected in an adaptive manner.
The dependence usually leads to bias in the sample average. 
This phenomenon is empirically documented in \cite{xu2013estimation,villar2015multi}.
Theoretically, \cite{nie2018adaptively} give sufficient conditions under which the sample mean is negative. 
\cite{shin2019sample,shin2019bias} provide a thorough analysis of the magnitude of the bias in multi-armed bandit problems.

How to debias the sample mean and learn the actual mean of the reward has attracted attention in recent literature. 
\cite{xu2013estimation} proposes a debiasing method by collecting another data log or a ``held-out'' set.
Collecting additional data improves the estimation in most cases, but it is not always feasible in many applications such as clinical trials.
In comparison, our method only relies on the bandit experiment without additional data.
In \cite{nie2018adaptively}, the authors provide an MLE-based approach that models the whole stochastic process and find the parameter (mean) that yields the maximum likelihood. 
Their approach requires the knowledge of the reward distribution to construct the likelihood function. \cite{neel2018mitigating}  shows that the bias
problem is mitigated if the data collection procedure is differentially private.
\cite{deshpande2018accurate} introduces a  ridge-type debiased estimator of the ordinary least squares estimator, which is computed adaptively to optimize bias-variance tradeoff. 
The debiasing procedure requires access to the data collection policy to construct a reasonable regularization parameter of the ridge estimation. 
Another line of literature uses the framework of causal inference. 
In particular, important sampling or propensity scoring has been used in  \cite{li2015toward,swaminathan2015counterfactual,schnabel2016recommendations,dimakopoulou2017estimation,vlassis2019design,hadad2021confidence} to develop off-policy evaluation schemes.
In order to evaluate the propensity score, the method typically only works for bandit algorithms with internal randomization such as Thompson sampling.

Our work builds on the bootstrap method, introduced by \cite{eforn1979bootstrap}.
Due to the data-dependent nature, bootstrap usually doesn't require any knowledge of the distribution itself and is well suited for multi-armed bandit problems.
For example, \cite{hao2019bootstrapping} replaces the confidence bound in UCB with the bootstrap version to circumvent the dependence on model parameters such as the subgaussian parameter of the reward distribution. 
\cite{eckles2014thompson,osband2015bootstrapped} use bootstrap in the posterior distribution of Thompson sampling to improve the computational efficiency. 
In addition, the bootstrap can be used to learn model coefficients
in contextual bandits \cite{tang2015personalized}, achieve near-optimal regret \cite{vaswani2018new}, approximate Thompson sampling \cite{elmachtoub2017practical} and conduct a generally well-performed algorithm in different models \cite{kveton2019garbage}.
These papers apply bootstrap as a component in the online algorithm, which is different from our work. 
Finally, bootstrap has also been used in the offline setting or off-policy evaluation.
\cite{kostrikov2020statistical,hao2021bootstrapping} propose a bootstrap Q-evaluation to evaluate the value of a target policy. 
\cite{bibaut2019more} obtains the confidence interval for its proposed doubly-robust estimator by bootstrapping. \cite{mary2014improving} uses 
bootstrapping techniques to improve 
offline evaluation of contextual bandit algorithms
in recommendation applications. 
In our paper, we focus on the bias of the sample average reward collected from a target multi-armed bandit algorithm and use bootstrap to simulate additional bandit experiments so that the bias can be evaluated.


\textbf{Notations.}
We use $K$ as the number of arms and $T$ as the number of rounds in the bandit experiment.
We use $\cdot^*$ for the corresponding quantity in the bootstrap world, conditional on the samples.

\section{Problem Formulation}
Suppose an agent has collected a dataset after running a bandit experiment.
In particular, the agent has observed the number of rounds $T$, the action sequence $\left\{a_1,\dots,a_T\right\}$ where $a_t\in [K]=\{1, \ldots, K\}$ is the arm pulled in round $t$, and the reward sequence $\left\{r_1,\dots,r_T\right\}$ where $r_t\in \R$ is the (noisy) reward observed in round $t$.
In addition, the agent is also aware of the bandit algorithm that has been implemented, i.e., the distribution of $a_{t}$ given $\Fscr_{t-1}\triangleq \sigma(a_1,r_1,\dots, a_{t-1}, r_{t-1})$. However, the reward distributions $P=\{P_1,\ldots,P_K\}$ are unknown, in particular, the mean $\mu_k$ for each arm $k$ is unknown.

The agent is interested in the average reward of, say, arm $k\in [K]$.
A straightforward approach is to take the sample average of the rewards generated from arm $k$: $\hat \mu_{k} = \frac{\sum_{t=1}^T r_t\I{a_t=k}}{\sum_{t=1}^T\I{a_t=k}}$.
As documented in the literature \cite{nie2018adaptively,shin2019sample,shin2019bias}, such a statistic is usually biased, especially for suboptimal arms. That is, the bias given by $\E_P[\hat \mu_{k}] - \mu_k$ is often negative, where $\mu_k$ is the mean of the reward of arm $k$.
However, the quantity depends on the algorithm and its interaction with the unknown $P$.
To show the intractability of this quantity, consider $\E_P[\sum_{t=1}^T\I{a_t=k}]$, the expected value of the total number of pulls of arm $k$ under a particular algorithm.
This seemingly simple quantity usually doesn't have a closed-form expression even when $P$ is known, for most algorithms such as UCB and Thompson sampling.
It is unclear if we can estimate the bias and use it to debias the sample averages.

Next, we propose a procedure to estimate the quantity based on bootstrap \cite{efron1994introduction} and in particular, multiplier bootstrap \cite{van1996weak,chernozhukov2013gaussian}.

\subsection{Multiplier Bootstrap}\label{sec:mb}
In this section, we first introduce the idea of multiplier bootstrap generally.
Let $\Xscr=\{z_1, \dots, z_n\}$ be i.i.d. observations from an unknown distribution $F$, whose mean is $\mu$.
Then the sample mean and sample variance can be calculated by $\hat{\mu}=\frac{z_1+\dots+z_n}{n}$ and $\hat{\sigma}^2=\frac{1}{n}\sum_{i=1}^{n}(z_i-\hat{\mu})^2$.
Conditional on $\Xscr$, the goal of bootstrap is to construct and simulate samples from $F$ approximately, say $\mathcal{X}^*=\{z_1^*, \dots, z_m^*\}$, without knowing $F$.

There are two bootstrap approaches involved in this study.
Efron's bootstrap (EB) \cite{efron1982jackknife} draws each sample $z_j^*$ uniformly randomly from $\Xscr$ with replacement.
That is $\PR(z_j^* = z_i) = \frac{1}{n}, \forall \: i=1,\dots,n$.
Equivalently, one can think of $z_j^* = \sum_{i=1}^n w_{ij}z_i$,
where the weight $(w_{1j}, \ldots, w_{nj})$ has a multinomial distribution with parameters $(1;1/n, \ldots, 1/n)$.

Another bootstrap approach, which is the focus of this paper, is referred to as multiplier bootstrap (MB) \cite{van1996weak}.
It has attracted the attention of many scholars due to its analytical tractability \cite{arlot2010some,chernozhukov2013gaussian,hao2019bootstrapping}.
In multiplier bootstrap, we have
$z_j^*=\frac{1}{\sqrt{n}}\sum_{i=1}^n(z_i-\hat{\mu})w_{ij}+\hat{\mu}$, where the weight $(w_{1j}, \ldots, w_{nj})$ is a random vector with $\E[w_{ij}]=0$ and $\Var[w_{ij}]=1$.
It is clear that
$\E[z_j^*]=\hat{\mu}$ and $\Var[z_j^*]=\hat{\sigma}^2$,
so MB preserves the sample mean and variance.

In our problem, the data $\Xscr$ consists of the action sequence $\left\{a_1,\dots,a_T\right\}$ and the reward sequence $\left\{r_{1},\dots,r_T\right\}$.
We use MB to construct the distribution of the reward of each arm with Gaussian weights.
Therefore, conditional on $\Xscr$, a reward from arm $k$ can be generated by
\begin{equation}\label{eq:mb-dist}
P^*_k\sim N(\hat\mu_k,\hat\sigma^2_k),
\end{equation}
where $\hat\mu_k$ and $\hat\sigma_k$ are the sample mean and variance of the observed rewards from arm $k$.

\subsection{Use Multiplier Bootstrap for Debiasing}\label{eq:mb-debias}
It is not a new idea to use bootstrap to correct the bias in the samples \cite{efron1994introduction, steck2003bias}.
But because of the structure of this application (the sequential nature of the observations), it is not straightforward to apply bootstrap.
We next elaborate on the major steps and intuition of the algorithm.
After collecting the data from the experiment, we simulate $B$ additional experiments in the bootstrap world, using the same $K$, $T$, and the bandit algorithm.
The reward of the bootstrap experiments is generated according to \eqref{eq:mb-dist}.
Comparing the experiments in the bootstrap and the real world, the bootstrap reward distribution only offers an approximation.
However, because the implemented bandit algorithm is known, the bootstrap experiments correctly capture the intertemporal dependence introduced by the sequential decision-making.
Therefore, the bias calculated in the bootstrap world may be used to estimate the bias in the real world, thanks to the known reward distribution in the bootstrap world.
Algorithm~\ref{alg:mb-debias} demonstrates the process.
\begin{algorithm}
    \caption{Using MB to debias the samples from a bandit experiment}
    \label{alg:mb-debias}
    \begin{algorithmic}[1]
        \Require The horizon $T$, the number of arms $K$, the action sequence $\left\{a_1,\dots,a_T\right\}$, the reward sequence $\left\{r_{1},\dots,r_T\right\}$,
        the implemented bandit algorithm
        \State Parameters: bootstrap sample size $B$
        \For{$b=1,\dots,B$}
        \State Simulate a bandit experiment with the same $K$, $T$ and bandit algorithm in the bootstrap world, where the reward of arm $k$ is generated by \eqref{eq:mb-dist}
        \State Record the action sequence $\left\{a^{*}_{b,1},\dots,a^{*}_{b,T}\right\}$ and the reward sequence $\left\{r^{*}_{b,1},\dots,r^{*}_{b,T}\right\}$
        \State Calculate the sample average of the reward of the arms in the experiment
        \begin{equation*}
            \hat\mu^*_{b,k} = \frac{\sum_{t=1}^T r^*_{b,t}\I{a^*_{b,t}=k}}{\sum_{t=1}^T\I{a^*_{b,t}=k}}, \quad \forall \:k=1,\dots,K
        \end{equation*}
        \EndFor
        \State Calculate the average bias in the bootstrap world $ \frac{1}{B}\sum_{b=1}^B\hat \mu^*_{b,k}-\hat\mu_k$ \label{step:bootstrap-bias}
        \State \Return the debiased sample average $2\hat \mu_k-\frac{1}{B}\sum_{b=1}^B\hat \mu^*_{b,k}$ as the estimator for $\mu_k$, $\forall \: k=1,\dots,K$\label{step:return}
    \end{algorithmic}
\end{algorithm}

\section{Numerical Performance}\label{sec:numerical}
In this section, we test the performance of Algorithm~\ref{alg:mb-debias} for bandit experiments using four popular bandit algorithms, including Explore-Then-Commit (ETC), upper confidence bound (UCB),
Thompson sampling (TS) and $\epsilon$-greedy (EG), for $T=100$ and $K=2$ (also $K=4$).
We consider two reward distributions: unit-variance normal distribution with $\mu_1=1$, $\mu_2=1.5$ and Bernoulli random variables with $\mu_1=0.3, \mu_2=0.6$.
The detailed implementations of the algorithms are given below.
\begin{itemize}
	\item ETC: pull each arm $m=10$ times sequentially in the first $mK$ rounds. Choose the arm with the highest average reward for the rest of the horizon.
	\item UCB: we use the version in Chapter 2.7 of \cite{sutton2018reinforcement}.
	In particular, in round $t$, the arm that maximizes $\hat \mu_k(t)+\sqrt{\log t/N_k(t)}$ is pulled, where $\hat \mu_k(t)$ is the empirical average reward of arm $k$ and $N_k(t)$ is the number of rounds arm $k$ is pulled prior to $t$.
	\item TS: pull the arm according to its posterior probability of being the optimal arm \cite{russo2017tutorial}. We set the prior distribution to be a normal distribution. 
	\item EG: pull the arm with the highest empirical average reward so far (greedy) with probability $1-\epsilon$ and selects a uniformly random arm with probability $\epsilon$. We set $\epsilon=0.05$. 
\end{itemize}
We consider $B=1,000$ bootstrap simulations in Algorithm~\ref{alg:mb-debias} when calculating the debiased sample average in Step~\ref{step:return}.
To evaluate the bias in the real world, $\E[\hat\mu_k]-\mu_k$, we run the bandit experiment 1,000 times and take the average.
We also report the debiased sample average using MB or EB. 
The numerical experiments are conducted on a PC with 3.10 GHz Intel Processor and 16 GB of RAM.

Moreover, we implement the two propensity score based methods in \cite{hadad2021confidence}: the inverse propensity score weighted (IPW) estimator and the augmented inverse propensity weighted (AIPW) estimator, for EG and TS.
Note that both estimators are unbiased: if we run a large number of bandit experiments and take the average bias, it is going to be zero. In addition, both estimators only work for bandit algorithms with internal randomization.
Therefore, we compare the mean-squared error (MSE) of our method and these two methods for EG and TS.


Table~\ref{table:two_arm} demonstrates the performance of Algorithm~\ref{alg:mb-debias}.
In all the instances, Algorithm~\ref{alg:mb-debias} is able to give a more accurate estimator for the mean reward of each arm. Although there are occasions when the bias is already quite small and Algorithm~\ref{alg:mb-debias} over-corrects, Algorithm~\ref{alg:mb-debias} is very effective in achieving bias reduction overall.
\begin{table}[h!]
	\centering
	\begin{tabular}{cccccccccc}
		\toprule
		\multirow{3}{*}{Algorithm} & \multirow{3}{*}{Reward}& \multicolumn{4}{c}{Real World} &\multicolumn{4}{c}{Bootstrap World}\\
		&&\multicolumn{2}{c}{$\E[\hat\mu_k]-\mu_k$} &\multicolumn{2}{c}{$\E[\hat{\mu}_k]$} &\multicolumn{2}{c}{Estimated Bias} & \multicolumn{2}{c}{Corrected $\E[\hat{\mu}_k]$}\\\cmidrule(lr){3-6}\cmidrule(lr){7-10}
		&&Arm 1 &Arm 2 &Arm 1 &Arm 2&Arm 1 &Arm 2&Arm 1& Arm 2\\\midrule
		\multirow{2}{*}{ETC} &Normal &-0.0518 &-0.0324 &0.9482 &1.4676 &-0.0377 &-0.0396 &0.9859 &1.5072\\
		&Bernoulli &-0.0184 &-0.0167 &0.2816 &0.5833 &-0.0148 &-0.0156 &0.2964 &0.5989\\\midrule
		\multirow{2}{*}{UCB} & Normal &-0.3407 &-0.0348 &0.6593 &1.4652 &-0.2026 &-0.0361 &0.8619 &1.5012\\
		&Bernoulli &-0.0384 &-0.0035 &0.2616 &0.5965 &-0.0406 &-0.003 &0.3022 &0.5996\\\midrule
		\multirow{2}{*}{TS} & Normal &-0.3181 &-0.0495 &0.6819 &1.4505 &-0.2602 &-0.0451 &0.9421 &1.4956\\
		&Bernoulli &-0.0309 &-0.0025 &0.2691 &0.5975 &-0.0451 &-0.0035 &0.3142 &0.6010\\\midrule
		\multirow{2}{*}{EG} &Normal &-0.167 &-0.109 &0.833 &1.391 &-0.1171 &-0.0958 &0.9501 &1.4868\\
		&Bernoulli &-0.0508 &-0.0828 &0.2492 &0.5172 &-0.0533 &-0.0377 &0.3025 &0.5549\\
		\bottomrule
	\end{tabular}
	\caption{Debaising using Algorithm~\ref{alg:mb-debias} for a two-armed bandit problem}\label{table:two_arm}
\end{table}

\begin{figure}[H]
	\centering
	\begin{tikzpicture}[scale=0.7]
	\begin{axis}[legend pos=north west,xlabel=Sample Mean,
	ymin=0, ymax = 800, xmin=0, xmax = 2, yticklabels={,,}]
	\addplot[hist={data=x},black, fill=black, fill opacity=0.1] table[x=arm1,ignore chars={\#},col sep=comma]{Data2/muhat11.dat};
	\addlegendentry{Arm1: $N(1,1)$}
	\addplot[hist={data=x},blue, fill=blue, fill opacity=0.1] table[x=arm2,ignore chars={\#},col sep=comma]{Data2/muhat11.dat};
	\addlegendentry{Arm2: $N(\frac{3}{2},1)$}
	\addplot[dashed,black] coordinates {(0.9482,0)(0.9482,10000)};
	\addplot[dashed,blue] coordinates {(1.4676,0)(1.4676,10000)};
	\addplot[red] coordinates {(1,0)(1,10000)};
	\addplot[red] coordinates {(1.5,0)(1.5,10000)};
	\end{axis}
	\end{tikzpicture}
	\hspace{0.01cm}
	\begin{tikzpicture}[scale=0.7]
	\begin{axis}[legend pos=north west,xlabel=MB Corrected Sample Mean,
	ymin=0, ymax = 800, xmin=0, xmax = 2, yticklabels={,,}]
	\addplot[hist={data=x},black,fill=black, fill opacity=0.1] table[x=arm1,ignore chars={\#},col sep=comma]{Data2/muhatmb11.dat};
	\addlegendentry{Arm1:$N(1,1)$}
	\addplot[hist={data=x},blue,fill=blue, fill opacity=0.1] table[x=arm2,ignore chars={\#},col sep=comma]{Data2/muhatmb11.dat};
	\addlegendentry{Arm2:$N(\frac{3}{2},1)$}
	\addplot[dashed,black] coordinates {(0.9859,0)(0.9859,10000)};
	\addplot[dashed,blue] coordinates {(1.5072,0)(1.5072,10000)};
	\addplot[red] coordinates {(1,0)(1,10000)};
	\addplot[red] coordinates {(1.5,0)(1.5,10000)};
	\end{axis} 
	\end{tikzpicture}
	
	\begin{tikzpicture}[scale=0.7]
	\begin{axis}[legend pos=north west,xlabel=Sample Mean,
	ymin=0, ymax = 800, xmin=0, xmax = 0.8,xtick={0,0.3,0.6}, yticklabels={,,}]
	\addplot[hist={data=x},black,fill=black,fill opacity=0.1] table[x=arm1,ignore chars={\#},col sep=comma]{Data2/muhat12.dat};
	\addlegendentry{Arm1: $Ber(0.3)$}
	\addplot[hist={data=x},blue,fill=blue,fill opacity=0.1] table[x=arm2,ignore chars={\#},col sep=comma]{Data2/muhat12.dat};
	\addlegendentry{Arm2: $Ber(0.6)$}
	\addplot[dashed,black] coordinates {(0.2816,0)(0.2816,10000)};
	\addplot[dashed,blue] coordinates {(0.5833,0)(0.5833,10000)};
	\addplot[red] coordinates {(0.3,0)(0.3,10000)};
	\addplot[red] coordinates {(0.6,0)(0.6,10000)};
	\end{axis}
	\end{tikzpicture}
	\hspace{0.01cm}
	\begin{tikzpicture}[scale=0.7]
	\begin{axis}[legend pos=north west,xlabel=MB Corrected Sample Mean,
	ymin=0, ymax = 800, xmin=0, xmax = 0.8,xtick={0,0.3,0.6}, yticklabels={,,}]
	\addplot[hist={data=x},black,fill=black,fill opacity=0.1] table[x=arm1,ignore chars={\#},col sep=comma]{Data2/muhatmb12.dat};
	\addlegendentry{Arm1:$Ber(0.3)$}
	\addplot[hist={data=x},blue,fill=blue,fill opacity=0.1] table[x=arm2,ignore chars={\#},col sep=comma]{Data2/muhatmb12.dat};
	\addlegendentry{Arm2: $Ber(0.6)$}
	\addplot[dashed,black] coordinates {(0.2964,0)(0.2964,10000)};
	\addplot[dashed,blue] coordinates {(0.5989,0)(0.5989,10000)};
	\addplot[red] coordinates {(0.3,0)(0.3,10000)};
	\addplot[red] coordinates {(0.6,0)(0.6,10000)};
	\end{axis} 
	\end{tikzpicture}
	\vspace{-0.3cm}
	\caption{Debiasing using bootstrap under ETC}
	\label{fig:etc}
\end{figure}
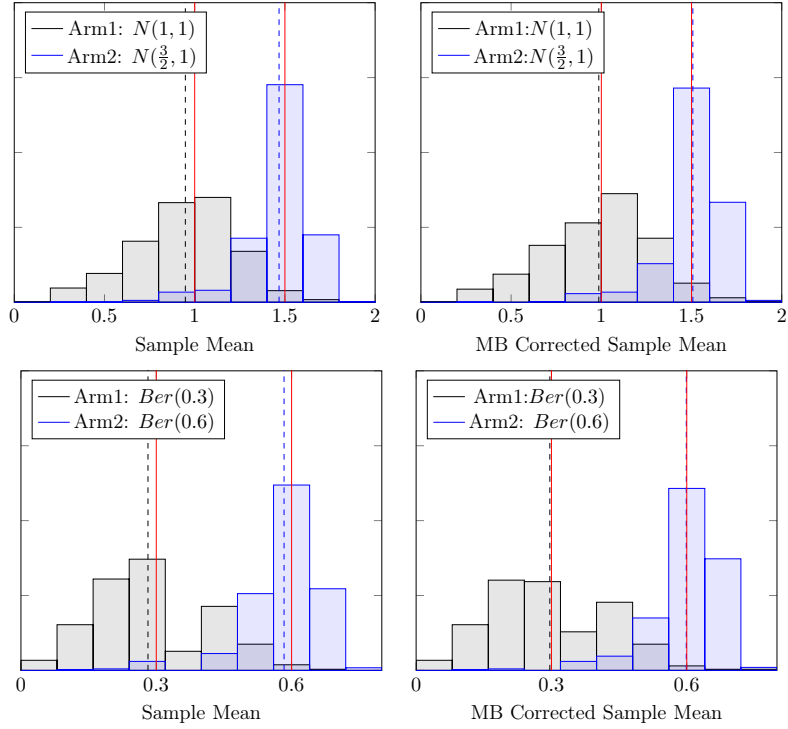

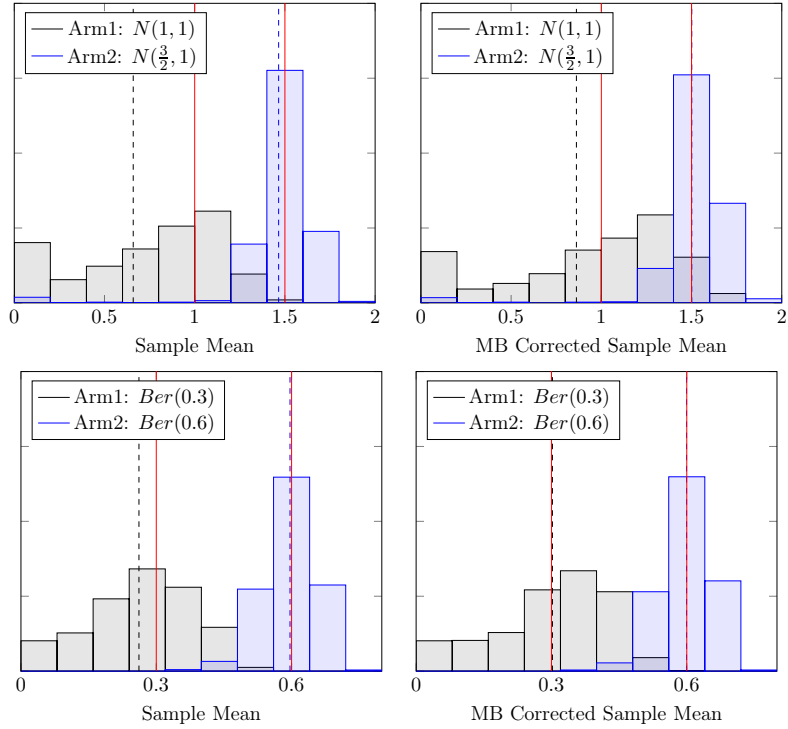
\begin{figure}[H]
	\centering
	\begin{tikzpicture}[scale=0.7]
	\begin{axis}[legend pos=north west,xlabel=Sample Mean,
	ymin=0, ymax = 800, xmin=0, xmax = 2, yticklabels={,,}]
	\addplot[hist={data=x},black,fill=black,fill opacity=0.1] table[x=arm1,ignore chars={\#},col sep=comma]{Data2/muhat21.dat};
	\addlegendentry{Arm1: $N(1,1)$}
	\addplot[hist={data=x},blue,fill=blue,fill opacity=0.1] table[x=arm2,ignore chars={\#},col sep=comma]{Data2/muhat21.dat};
	\addlegendentry{Arm2: $N(\frac{3}{2},1)$}
	\addplot[dashed,black] coordinates {(0.6593,0)(0.6593,10000)};
	\addplot[dashed,blue] coordinates {(1.4652,0)(1.4652,10000)};
	\addplot[red] coordinates {(1,0)(1,10000)};
	\addplot[red] coordinates {(1.5,0)(1.5,10000)};
	\end{axis}
	\end{tikzpicture}
	\hspace{0.01cm}
	\begin{tikzpicture}[scale=0.7]
	\begin{axis}[legend pos=north west,xlabel=MB Corrected Sample Mean,
	ymin=0, ymax = 800, xmin=0, xmax = 2, yticklabels={,,}]
	\addplot[hist={data=x},black,fill=black,fill opacity=0.1] table[x=arm1,ignore chars={\#},col sep=comma]{Data2/muhatmb21.dat};
	\addlegendentry{Arm1: $N(1,1)$}
	\addplot[hist={data=x},blue,fill=blue,fill opacity=0.1] table[x=arm2,ignore chars={\#},col sep=comma]{Data2/muhatmb21.dat};
	\addlegendentry{Arm2: $N(\frac{3}{2},1)$}
	\addplot[dashed,black] coordinates {(0.8619,0)(0.8619,10000)};
	\addplot[dashed,blue] coordinates {(1.5012,0)(1.5012,10000)};
	\addplot[red] coordinates {(1,0)(1,10000)};
	\addplot[red] coordinates {(1.5,0)(1.5,10000)};
	\end{axis} 
	\end{tikzpicture}
	
	\begin{tikzpicture}[scale=0.7]
	\begin{axis}[legend pos=north west,xlabel=Sample Mean,
	ymin=0, ymax = 800, xmin=0, xmax = 0.8,xtick={0,0.3,0.6}, yticklabels={,,}]
	\addplot[hist={data=x},black,fill=black,fill opacity=0.1] table[x=arm1,ignore chars={\#},col sep=comma]{Data2/muhat22.dat};
	\addlegendentry{Arm1: $Ber(0.3)$}
	\addplot[hist={data=x},blue,fill=blue,fill opacity=0.1] table[x=arm2,ignore chars={\#},col sep=comma]{Data2/muhat22.dat};
	\addlegendentry{Arm2: $Ber(0.6)$}
	\addplot[dashed,black] coordinates {(0.2616,0)(0.2616,10000)};
	\addplot[dashed,blue] coordinates {(0.5965,0)(0.5965,10000)};
	\addplot[red] coordinates {(0.3,0)(0.3,10000)};
	\addplot[red] coordinates {(0.6,0)(0.6,10000)};
	\end{axis}
	\end{tikzpicture}
	\hspace{0.01cm}
	\begin{tikzpicture}[scale=0.7]
	\begin{axis}[legend pos=north west,xlabel=MB Corrected Sample Mean,
	ymin=0, ymax = 800, xmin=0, xmax = 0.8,xtick={0,0.3,0.6}, yticklabels={,,}]
	\addplot[hist={data=x},black,fill=black,fill opacity=0.1] table[x=arm1,ignore chars={\#},col sep=comma]{Data2/muhatmb22.dat};
	\addlegendentry{Arm1: $Ber(0.3)$}
	\addplot[hist={data=x},blue,fill=blue,fill opacity=0.1] table[x=arm2,ignore chars={\#},col sep=comma]{Data2/muhatmb22.dat};
	\addlegendentry{Arm2: $Ber(0.6)$}
	\addplot[dashed,black] coordinates {(0.3022,0)(0.3022,10000)};
	\addplot[dashed,blue] coordinates {(0.5996,0)(0.5996,10000)};
	\addplot[red] coordinates {(0.3,0)(0.3,10000)};
	\addplot[red] coordinates {(0.6,0)(0.6,10000)};
	\end{axis} 
	\end{tikzpicture}
	\vspace{-0.3cm}
	\caption{Debiasing using bootstrap under UCB}
	\label{fig:ucb}
\end{figure}

In Figures~\ref{fig:etc} and~\ref{fig:ucb}, we provide more information about the performance of Algorithm~\ref{alg:mb-debias}.
In particular, for ETC, we show the histogram of the raw sample mean as well as MB corrected sample means among the 1,000 bandit experiments.
The average is illustrated by the dashed vertical lines and the actual mean reward by the solid vertical lines. 
Again, after applying MB to the sample means, the bias is significantly reduced.

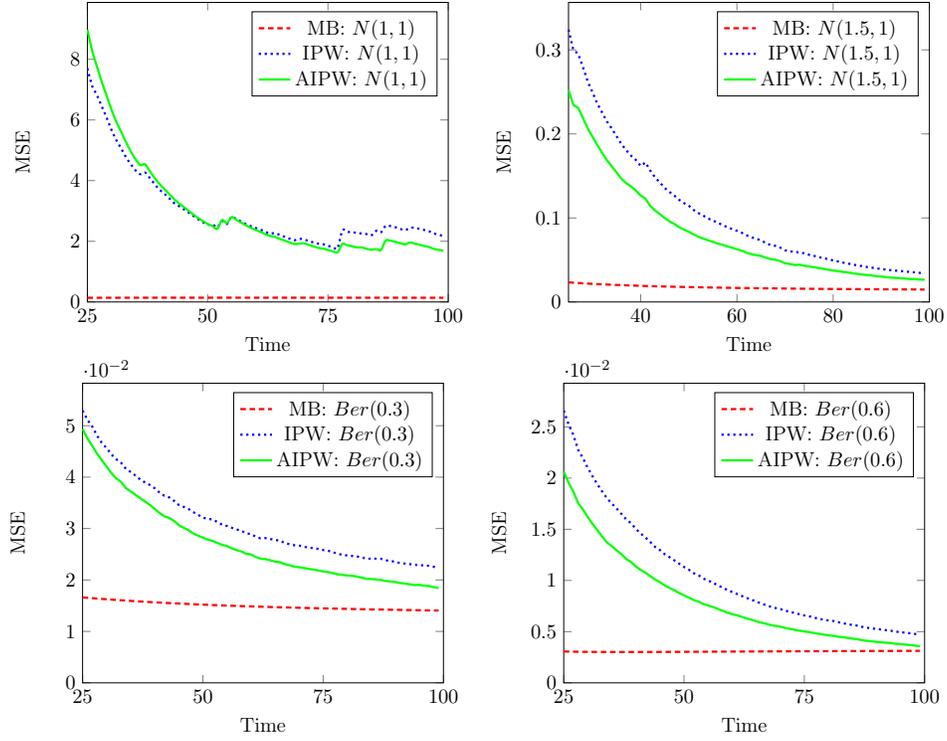
\begin{figure}[H]
	\centering
	\begin{tikzpicture}[scale=0.7]
	\begin{axis}[legend pos=north east,xlabel=Time,ylabel=MSE, ymin=0,xmin=25, xmax = 100,xtick={25,50,75,100}]
	\addplot[smooth,color=red, densely dashed, very thick] table[x=time,y=arm1,ignore chars={\#},col sep=comma] {MSE/MSEmb31.dat};
	\addlegendentry{MB: $N(1,1)$}
	\addplot[smooth,color=blue,dotted,very thick] table[x=time,y=arm1,ignore chars={\#},col sep=comma]{MSE/MSEipw31.dat};	
	\addlegendentry{IPW: $N(1,1)$}
	\addplot[smooth,color=green,very thick] table[x=time,y=arm1,ignore chars={\#},col sep=comma]{MSE/MSEaipw31.dat};		
	\addlegendentry{AIPW: $N(1,1)$}
	\end{axis}
	\end{tikzpicture}
	\hspace{0.01cm}
	\begin{tikzpicture}[scale=0.7]
	\begin{axis}[legend pos=north east,xlabel=Time,ylabel=MSE,ymin=0,xmin=25, xmax = 100]
	\addplot[color=red,densely dashed, very thick] table[x=time,y=arm2,ignore chars={\#},col sep=comma] {MSE/MSEmb31.dat};
	\addlegendentry{MB: $N(1.5,1)$}
	\addplot[color=blue,dotted,very thick] table[x=time,y=arm2,ignore chars={\#},col sep=comma]{MSE/MSEipw31.dat};	
	\addlegendentry{IPW: $N(1.5,1)$}
	\addplot[color=green,very thick] table[x=time,y=arm2,ignore chars={\#},col sep=comma]{MSE/MSEaipw31.dat};		
	\addlegendentry{AIPW: $N(1.5,1)$}
	\end{axis}
	\end{tikzpicture}
	
	\begin{tikzpicture}[scale=0.7]
	\begin{axis}[legend pos=north east,xlabel=Time,ylabel=MSE,ymin=0,xmin=25, xmax = 100,xtick={25,50,75,100}]
	\addplot[color=red,densely dashed, very thick] table[x=time,y=arm1,ignore chars={\#},col sep=comma] {MSE/MSEmb32.dat};
	\addlegendentry{MB: $Ber(0.3)$}
	\addplot[color=blue, dotted, very thick] table[x=time,y=arm1,ignore chars={\#},col sep=comma]{MSE/MSEipw32.dat};	
	\addlegendentry{IPW: $Ber(0.3)$}
	\addplot[color=green,very thick] table[x=time,y=arm1,ignore chars={\#},col sep=comma]{MSE/MSEaipw32.dat};		
	\addlegendentry{AIPW: $Ber(0.3)$}
	\end{axis}
	\end{tikzpicture}
	\hspace{0.01cm}
	\begin{tikzpicture}[scale=0.7]
	\begin{axis}[legend pos=north east,xlabel=Time,ylabel=MSE,ymin=0,xmin=25, xmax = 100,xtick={25,50,75,100}]
	\addplot[color=red,densely dashed, very thick] table[x=time,y=arm2,ignore chars={\#},col sep=comma] {MSE/MSEmb32.dat};
	\addlegendentry{MB: $Ber(0.6)$}
	\addplot[color=blue,dotted, very thick] table[x=time,y=arm2,ignore chars={\#},col sep=comma]{MSE/MSEipw32.dat};	
	\addlegendentry{IPW: $Ber(0.6)$}
	\addplot[color=green,very thick] table[x=time,y=arm2,ignore chars={\#},col sep=comma]{MSE/MSEaipw32.dat};		
	\addlegendentry{AIPW: $Ber(0.6)$}
	\end{axis}
	\end{tikzpicture}
	\vspace{-0.3cm}
	\caption{MSE comparison under TS}
	\label{MSE_TS}
\end{figure}

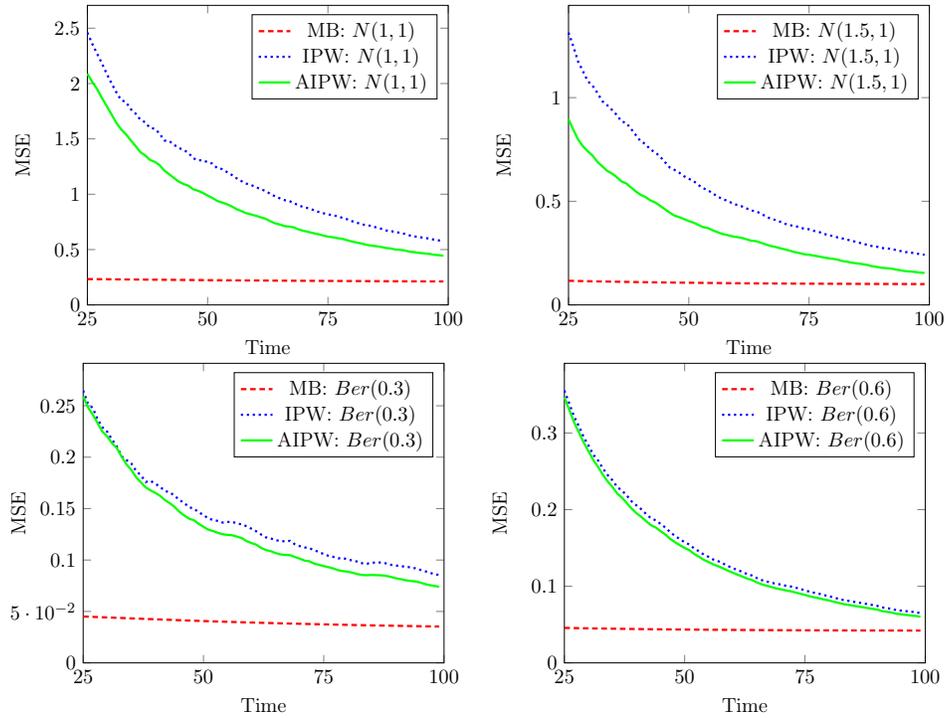
\begin{figure}[H]
	\centering
	\begin{tikzpicture}[scale=0.7]
	\begin{axis}[legend pos=north east,xlabel=Time,ylabel=MSE, ymin=0,xmin=25, xmax = 100,xtick={25,50,75,100}]
	\addplot[color=red,densely dashed, very thick] table[x=time,y=arm1,ignore chars={\#},col sep=comma] {MSE/MSEmb41.dat};
	\addlegendentry{MB: $N(1,1)$}
	\addplot[color=blue,dotted,very thick] table[x=time,y=arm1,ignore chars={\#},col sep=comma]{MSE/MSEipw41.dat};	
	\addlegendentry{IPW: $N(1,1)$}
	\addplot[color=green,very thick] table[x=time,y=arm1,ignore chars={\#},col sep=comma]{MSE/MSEaipw41.dat};		
	\addlegendentry{AIPW: $N(1,1)$}
	\end{axis}
	\end{tikzpicture}
	\hspace{0.01cm}
	\begin{tikzpicture}[scale=0.7]
	\begin{axis}[legend pos=north east,xlabel=Time,ylabel=MSE,ymin=0,xmin=25, xmax = 100,xtick={25,50,75,100}]
	\addplot[color=red,densely dashed, very thick] table[x=time,y=arm2,ignore chars={\#},col sep=comma] {MSE/MSEmb41.dat};
	\addlegendentry{MB: $N(1.5,1)$}
	\addplot[color=blue,dotted,very thick] table[x=time,y=arm2,ignore chars={\#},col sep=comma]{MSE/MSEipw41.dat};	
	\addlegendentry{IPW: $N(1.5,1)$}
	\addplot[color=green,very thick] table[x=time,y=arm2,ignore chars={\#},col sep=comma]{MSE/MSEaipw41.dat};		
	\addlegendentry{AIPW: $N(1.5,1)$}
	\end{axis}
	\end{tikzpicture}
	
	\begin{tikzpicture}[scale=0.7]
	\begin{axis}[legend pos=north east,xlabel=Time,ylabel=MSE,ymin=0,xmin=25, xmax = 100,xtick={25,50,75,100}]
	\addplot[color=red,densely dashed, very thick] table[x=time,y=arm1,ignore chars={\#},col sep=comma] {MSE/MSEmb42.dat};
	\addlegendentry{MB: $Ber(0.3)$}
	\addplot[color=blue,dotted,very thick] table[x=time,y=arm1,ignore chars={\#},col sep=comma]{MSE/MSEipw42.dat};	
	\addlegendentry{IPW: $Ber(0.3)$}
	\addplot[color=green,very thick] table[x=time,y=arm1,ignore chars={\#},col sep=comma]{MSE/MSEaipw42.dat};		
	\addlegendentry{AIPW: $Ber(0.3)$}
	\end{axis}
	\end{tikzpicture}
	\hspace{0.01cm}
	\begin{tikzpicture}[scale=0.7]
	\begin{axis}[legend pos=north east,xlabel=Time,ylabel=MSE,ymin=0,xmin=25, xmax = 100,xtick={25,50,75,100}]
	\addplot[color=red,densely dashed, very thick] table[x=time,y=arm2,ignore chars={\#},col sep=comma] {MSE/MSEmb42.dat};
	\addlegendentry{MB: $Ber(0.6)$}
	\addplot[color=blue,dotted,very thick] table[x=time,y=arm2,ignore chars={\#},col sep=comma]{MSE/MSEipw42.dat};	
	\addlegendentry{IPW: $Ber(0.6)$}
	\addplot[color=green,very thick] table[x=time,y=arm2,ignore chars={\#},col sep=comma]{MSE/MSEaipw42.dat};		
	\addlegendentry{AIPW: $Ber(0.6)$}
	\end{axis}
	\end{tikzpicture}
	\vspace{-0.3cm}
	\caption{MSE comparison under EG}
	\label{MSE_EG}
\end{figure}

To compare to IPW and AIPW, Figures~\ref{MSE_TS} and~\ref{MSE_EG} illustrate the MSE of the corrected mean rewards among the 1000 bandit experiments. 
When $T=100$, all methods achieve a similar level of MSE. 
When $T$ is small, however, the MSE of IPW and AIPW is much larger. 
A large variance is a common issue for propensity score based methods.
By trading off a little bias for variance reduction, Algorithm~\ref{alg:mb-debias} seems to be more robust for small $T$.

We also conduct additional bandit experiments with $K=4$ arms:
Gaussian reward with mean $\mu=\{2,2.5,3,3.5\}$ and standard deviation $\sigma=\{2,1,2,1\}$, and Bernoulli reward with mean $\mu=\{0.4, 0.5, 0.7, 0.8\}$.
The four algorithms are repeated $1,000$ times, each instance with $B=1,000$ bootstrap simulations for both Efron's bootstrap and Gaussian multiplier bootstrap.
The results are presented in Appendix~\ref{sec:four-armed}.
They further demonstrate the effectiveness of the proposed algorithm.

\section{Theoretical Results for ETC}\label{sec:ETCtheory}
In this section, we analyze the performance of Algorithm~\ref{alg:mb-debias} when the ETC algorithm (see Section~\ref{sec:numerical}) is used in the bandit experiment.
The next lemma characterizes the sign of the bias of the ETC algorithm in the real world.

\begin{lemma}\label{lem:bias_negative}
Bias of arm $k$, $\E_P[\hat \mu_{k}] - \mu_k$, is negative when the ETC algorithm is used for all $k \in [K]$.
\end{lemma}
Although the result is a special case of \cite{nie2018adaptively}, in the proof (see Appendix~\ref{sec:lemma1}) we are able to explicitly characterize the bias of the sample average of arm $k$, in the form 
\begin{align} \label{eq:bias-etc}
\E_P[\hat \mu_{k}] - \mu_k 
 =   \frac{T-mK}{m + T - m K} \cdot \E_P \left[ ( \mu_k  - \hat \mu_{k}(mK) ) \I{a_{mK+1}=k} \right],
\end{align}
where $\hat \mu_{k}(mK)$ is the sample average reward of arm $k$ in the first $mK$ rounds.
The bias is fully generated by the correlation between $\mu_k-\hat \mu_k(mK)$, which is the deviation of the sample average in the exploration phase from the mean,
and $\I{a_{mK+1}=k}$, which is the event that arm $k$ is chosen in the exploitation phase. 
The expression \eqref{eq:bias-etc} allows us to derive the main theoretical results in the paper.

\subsection{Two Arms with Gaussian Rewards}\label{sec:two-arm-Gaussian}
In this section, we consider the special case when $K=2$ and the rewards of both arms are generated from the normal distribution.
That is, the reward of arm $i$ is generated from
$N(\mu_i, \sigma_i^2)$ for $i=1,2$.
We first compute the bias explicitly using the sample mean when the ETC algorithm is used.
\begin{proposition}\label{prop:two_normal_true_bias}
Under the ETC algorithm, for $K=2$ and Gaussian rewards, the bias of the sample average of arm $k$'s reward is
\begin{align} \label{eq:bias-2arm-normal}
    \E_P[\hat \mu_k] - \mu_k=-\frac{T-2 m}{ T - m }\frac{\sigma
    _k^2}{\sqrt{2\pi(\sigma_1^2+\sigma_2^2)m}}\exp \left( -\frac{m}{2(\sigma_1^2+\sigma_2^2)} (\mu_1-\mu_2)^2\right).
\end{align}
\end{proposition}
We have the following observations from Proposition~\ref{prop:two_normal_true_bias}:
\begin{itemize}
    \item The bias is always negative for both arms, confirming Lemma~\ref{lem:bias_negative}.
    \item If $\sigma_1=\sigma_2$, then the bias is equal for both arms.
        This may be counter-intuitive, as one would expect that the better arm is more likely to be chosen after the exploration phase and the sample average is thus less biased.
    \item Given $\sigma_1$ and $\sigma_2$, the bias is maximized when $\mu_1=\mu_2$, i.e., when the two arms are hard to distinguish.
    \item Given the reward distributions, the bias is decreasing exponentially in $m$.
\end{itemize}

To prove Proposition~\ref{prop:two_normal_true_bias}, the main idea is to use the bias characterization \eqref{eq:bias-etc}. Under the Gaussian assumption of reward distributions, the sample average $\hat \mu_k(mK)$ in the exploration is still Gaussian; In addition, whether the arm $k$ is chosen in the exploitation phase amounts to the comparison of average rewards of two arms which both follow Gaussian distributions. This allows us to obtain the explicit formula for the bias in \eqref{eq:bias-2arm-normal} in the two-arm setting. 
For multiple arms, it is difficult to obtain such explicit formulas due to the comparison of multiple Gaussian distributions  \cite{kim2007recent}.   

Next, we investigate the effect of using Gaussian multiplier bootstrap to estimate \eqref{eq:bias-2arm-normal} and correct the bias.
In order to generate the reward in the bootstrap world, we need to obtain the sample mean and variance in \eqref{eq:mb-dist}.
There are two outcomes in the context of ETC.
If arm one has a higher average in the exploration phase, then we have 
\begin{align*}
    \hat{\mu}_1&=\frac{1}{T-m}\left(\sum_{i=1}^m r_i +\sum_{i=2m+1}^T r_i\right), \; &\hat{\sigma}_1^2&=\frac{1}{T-m}\left(\sum_{i=1}^m (r_i-\hat \mu_1)^2 +\sum_{i=2m+1}^T (r_i-\hat\mu_1)^2\right),\\
    \hat{\mu}_2&=\frac{1}{m}\sum_{i=m+1}^{2m} r_i , \;&\hat{\sigma}_2^2&=\frac{1}{m}\sum_{i=m+1}^{2m}(r_i-\hat{\mu}_2)^2.
\end{align*}
This outcome happens with probability $\mathbb{P}(N(\mu_1,\frac{\sigma_1^2}{m})>N(\mu_2,\frac{\sigma_2^2}{m}))=\Phi\left(\frac{(\mu_1-\mu_2)\sqrt{m}}{\sqrt{\sigma_1^2+\sigma_2^2}}\right)$.
Otherwise, arm two has a higher average in the exploration phase.
We have
\begin{align*}
    \hat{\mu}_1&=\frac{1}{m}\sum_{i=1}^m r_i , \;&\hat{\sigma}_1^2&=\frac{1}{m}\sum_{i=1}^{m}(r_i-\hat{\mu}_1)^2,\\
    \hat{\mu}_2&=\frac{1}{T-m}\sum_{i=m+1}^T r_i, \; &\hat{\sigma}_2^2&=\frac{1}{T-m}\sum_{i=m+1}^T (r_i-\hat \mu_2)^2.
\end{align*}
and this outcome occurs with probability $\Phi\left(\frac{(\mu_2-\mu_1)\sqrt{m}}{\sqrt{\sigma_1^2+\sigma_2^2}}\right)$.

After obtaining $\hat\mu_k$ and $\hat\sigma_k$ for $k=1,2$, we can apply Proposition~\ref{prop:two_normal_true_bias} to the bandit experiments in the bootstrap world, i.e., Step~\ref{step:bootstrap-bias} in Algorithm~\ref{alg:mb-debias}.
In particular, when $B\to\infty$, the law of large numbers implies that the bias in the bootstrap world is given by
\begin{align}
    &\lim_{B\to\infty}\frac{1}{B}\sum_{b=1}^B\hat \mu^*_{b,k}-\hat\mu_k\\
    &= \E_{P^*}[\hat \mu_k^*] - \hat{\mu}_k\notag \\
    &=-\frac{T-2m}{ T - m } \frac{\hat{\sigma}_k^2}{\sqrt{2\pi(\hat{\sigma}_1^2+\hat{\sigma}_2^2)m}}\exp \left[ -\frac{m}{2(\hat{\sigma}_1^2+\hat{\sigma}_2^2)} (\hat{\mu}_1-\hat{\mu}_2)^2\right]. \label{eq:MB-bias}
\end{align}
Note that the bias is conditional on the data obtained from bandit experiment in the real world, $\Xscr$, \emph{i.e.,} the action sequence $\left\{a_1,\dots,a_T\right\}$, the reward sequence $\left\{r_{1},\dots,r_T\right\}$.

To compare the estimated bias in the bootstrap world, \eqref{eq:MB-bias}, to the actual bias in the real world, \eqref{eq:bias-2arm-normal},
note that in the bootstrap world, we simply replace $\mu_k$ and $\sigma_k$ by the sample version $\hat\mu_k$ and $\hat\sigma_k$.
Such analytical tractability is one of the major benefits of Gaussian multiplier bootstrap.
We define following quantity:
\begin{align}
    g_k(\mu_1, \mu_2, \sigma_1^2, \sigma_2^2) = \log \left( \frac{T-2 m}{ T - m }\frac{\sigma
    _k^2}{\sqrt{2\pi(\sigma_1^2+\sigma_2^2)m}}\exp \left[ -\frac{m}{2(\sigma_1^2+\sigma_2^2)} (\mu_1-\mu_2)^2\right] \right),
\end{align}
which is the logarithm of the absolute value of bias for arm $k$ in the real world.
The corresponding quantity in the bootstrap world is thus $g_k(\hat\mu_1, \hat\mu_2, \hat\sigma_1^2, \hat\sigma_2^2)$.
We focus on the logarithm because, as shown in Proposition~\ref{prop:two_normal_true_bias}, the bias decays exponentially fast as $m$ increases.
Therefore, we focus on the ratio of the decay rate.
Our next result states that the ratio converges to one asymptotically.
\begin{theorem}\label{thm:decayrate_normal}
Consider the ETC algorithm for $K=2$ and Gaussian rewards.
Choosing $T > 2m$, we have
\begin{equation*}
    \lim_{m\to \infty}\frac{g_k(\hat{\mu}_1, \hat{\mu}_2, \hat{\sigma}_1^2, \hat{\sigma}_2^2) }{ g_k(\mu_1, \mu_2, \sigma_1^2, \sigma_2^2)}=1, \quad \text{in probability for $k=1,2$}.
\end{equation*}
\end{theorem}
Theorem~\ref{thm:decayrate_normal} provides a theoretical justification for the performance of Algorithm~\ref{alg:mb-debias}.
At least for the ETC algorithm, when the exploration phase lengthens, the biases from the bootstrap world and the real world converge in the logarithmic sense.

Instead of the ratio of the log-biases, one may hope to obtain a stronger result: the convergence of the ratio of the biases themselves.
We caution that the claim is not true in general.
This is because the exponential may distort a small error and lead to divergence.

To prove Theorem~\ref{thm:decayrate_normal}, we perform a (stochastic) Taylor expansion of $g_k(\hat{\mu}_1, \hat{\mu}_2, \hat{\sigma}_1^2, \hat{\sigma}_2^2)$ at the point $(\mu_1, \mu_2, \sigma_1^2, \sigma_2^2)$. We then show that $\nabla g_k/g_k = O(1)$ and 
$||(\hat{\mu}_1-\mu_1,
\hat{\mu}_2-\mu_2,
\hat{\sigma}_1^2-\sigma_1^2,
\hat{\sigma}_2^2-\sigma_2^2)||$ converges in probability to zero as $m \rightarrow \infty$. The details of the proof are given in Appendix~\ref{sec:thm1}.


\subsection{General Rewards}\label{sec:general-reward}
In this section, we extend the setting in Section~\ref{sec:two-arm-Gaussian} by considering general rewards.
In particular, we consider the ETC algorithm with $K=2$ and the subgaussian reward distributions.
The reward distribution of the first arm has mean $\mu_1$ and variance proxy parameter $a^2$, i.e. $\E[e^{\lambda (X_1 - \mu_1)}] \le \exp(\frac{\lambda^2 a^2}{2})$ for all $\lambda \in \mathbb{R}.$ It is well known that $\sigma_1^2\defeq \Var(X_1) \le a^2$, where the equality holds when $X$ is Gaussian.
For simplicity, we assume the reward of the second arm is deterministic and equals $\mu_2$.
The analysis can be extended to the general subgaussian setting.

{To facilitate the presentation, we recall some definitions from the large deviations theory; see, e.g., \cite{dembo1998large} for background. 
Define the log moment generating function $\eta (h) = \log  \E [e^{h X_1}]$ for the reward of arm one.
Note $\eta$ is convex and continuously differentiable on $\mathbb{R}$ with $\eta'(0)= \E[X_1]= \mu_1$ (Lemma 2.2.5 of \cite{dembo1998large}). Set $\mathcal{H}=\{\eta'(h): h \in \mathbb{R}\}$.
The Legendre-Fenchel transform of $\eta$, denoted as $\Lambda^*$, is given by
\begin{align}\label{eq:LF-rate}
\Lambda^*(x)= \sup\limits_{h \in \mathbb{R}} ( hx - \eta(h) ).
\end{align}
Define $\zeta $ by the implicit equation $\eta'(\zeta ) =\mu_2$ when $\mu_2 \in \mathcal{H}$.\footnote{The case of $\mu_2 \notin \mathcal{H}$ is not very interesting in terms of analysis. For instance if the reward $X_1$ follows a uniform distribution on $(a, b)$, then $\mathcal{H}=(a,b)$. If $\mu_2$ lies outside of this interval, say, $\mu_2>b,$ it means with probability one the reward of arm 1 is smaller than the reward of arm 2. So the ETC algorithm always selects arm 2 in the exploit/commit phase and the bias analysis is simpler.} 
The next result characterizes the bias in this setting in the real world.
\begin{proposition}\label{prop:two_subGaussian_true_bias}
Under the ETC algorithm, for $K=2$, subgaussian reward for arm one and deterministic reward for arm two, the bias of the sample average of arm one's reward is
\begin{align} \label{eq:decay-rate-general}
\E_P[\hat \mu_{1}] - \mu_1 =   \frac{T-2m}{T - m } \frac{e^{-m \Lambda^*(\mu_2)}}{ \sqrt{2 \pi m \eta''(\zeta )}}  (-c_* + o(1)),
\end{align}
as $m \rightarrow \infty$, where $c_*>0$ is a constant.
\end{proposition}
The constant $c_*$ in the expression is semi-explicit.
The value depends on whether the reward $X_1$ has a lattice or non-lattice law, which is given in \eqref{eq:c_*} in the appendix.
Proposition~\ref{prop:two_subGaussian_true_bias} shows that the bias of arm one decays exponentially fast to zero with rate $\Lambda^*(\mu_2)$ as $m$ grows.
This is a similar pattern to Proposition~\ref{prop:two_normal_true_bias}.
In fact, in the special case that $X_1 \sim N(\mu_1, \sigma_1^2)$, one can readily verify that $\eta(h)=\mu_1h+\frac{1}{2}\sigma_1^2h^2$, $\Lambda^*(\mu_2) =  \frac{1}{2{\sigma}_1^2} ({\mu}_1 - {\mu}_2)^2$ and $c_*=\sigma_1^2$.
Hence, \eqref{eq:decay-rate-general} is consistent with \eqref{eq:bias-2arm-normal} when the rewards are Gaussian.

To prove Proposition~\ref{prop:two_subGaussian_true_bias}, we build upon the bias characterization in \eqref{eq:bias-etc}, and use tools from large deviations theory. We need the large deviations theory since one has to compare the average rewards of two arms in the exploration to decide which arm to pull in the exploitation phase for the ETC algorithm.
Mathematically, to obtain \eqref{eq:decay-rate-general}, we build on the Bahadur-Rao theorem (Theorem 3.7.4 in \cite{dembo1998large})
and provide precise estimates for the
tail probabilities and tail expectations for the average of i.i.d.\ random variables.
The details of the proof can be found in Appendix~\ref{sec:subgaussian}. }

Next, we consider the bias estimate in the bootstrap world.
With Gaussian bootstrap multiplier, the bootstrap distribution \eqref{eq:mb-dist} is always Gaussian with distribution $N(\hat\mu_1,\hat\sigma^2_1)$, capturing the first- and second-order moments of the original distribution.
As a result, we can infer from \eqref{eq:MB-bias} that
the bias of arm one in the bootstrap world is given by
\begin{align} 
    \lim_{B\to\infty}\frac{1}{B}\sum_{b=1}^B\hat \mu^*_{b,1}-\hat\mu_1 &= \E_{P^*}[\hat \mu_1^*] - \hat{\mu}_1 =-\frac{T-2m}{ T - m } \frac{\hat{\sigma}_1^2}{\sqrt{2\pi \hat{\sigma}_1^2 m}}\exp \left[ -m  \hat \Lambda^*(\mu_2) \right],
\end{align}
where the exponential decay rate of the bias in the bootstrap world is given by
\begin{align} \label{eq:hat-LF}
  \hat \Lambda^*(\mu_2) =  \frac{1}{2\hat{\sigma}_1^2} (\hat{\mu}_1 - {\mu}_2)^2.
\end{align} 
Comparing $\hat\Lambda^*(\mu_2)$ in \eqref{eq:hat-LF} to $\Lambda^*(\mu_2)$ in \eqref{eq:decay-rate-general}, we have the following theorem
\begin{theorem}\label{thm:decayrate_subGaussian}
Under the same assumptions of Proposition~\ref{prop:two_subGaussian_true_bias}, 
the ratio of the decay rate of the biases in the bootstrap world and the real world is asymptotically bounded by
\begin{align} \label{eq:sub-bound}
   \lim_{m\to\infty} \frac{\hat\Lambda^*(\mu_2)}{\Lambda^*(\mu_2)} \leq \frac{a^2 }{\sigma_1^2},
\end{align}
where $\sigma_1^2$ and $a^2$ are the variance and the variance proxy of the subgaussian reward from arm one, respectively. 
\end{theorem}
Theorem~\ref{thm:decayrate_subGaussian} characterizes the relationship of the decay rates of the biases in the real and bootstrap worlds.
It reveals one potential drawback of bootstrap despite its good performance in Section~\ref{sec:numerical}.
Only the first- and second-order moments captured by the Gaussian multiplier bootstrap are not sufficient to characterize the asymptotic decay rate.
The actual decay rate may depend on the tail behavior, which is captured by $a$ and the Legendre-Fenchel transform $\Lambda^*$. 
This is in general hard to capture using bootstrap, including Efron's bootstrap.
Still, Theorem~\ref{thm:decayrate_subGaussian} provides a bound for the ratio, which is attained when the reward indeed has a Gaussian distribution.

To prove Theorem~\ref{thm:decayrate_subGaussian}, we note that one can expect $||\hat{\mu}_1 - \mu_1|| + || \hat{\sigma}_1^2 -\sigma_1^2|| {\rightarrow} 0$ as $m \rightarrow \infty$. On combining with the Chernoff-Hoeffding bound for the average of i.i.d. sub-gaussian rewards from arm one, we can obtain \eqref{eq:sub-bound}. 
The details of the proof can be found in Appendix~\ref{sec:proof-thm2}.

\section{Conclusions, Limitations and Future Research} \label{sec:conclusion}
In this paper, we study the problem of inferring the mean reward of the arms from the data generated in a multi-armed bandit experiment.
The procedure we propose is based on bootstrap and doesn't require any knowledge of the reward distribution.
It can be used for a wide range of bandit algorithms.
We demonstrate strong performance in numerical examples and analyze the theoretical properties for the ETC algorithm.

Our current theoretical analysis is limited to the ETC algorithm. It remains an open direction to analyze the procedure for other algorithms such as UCB and Thompson sampling. 
The major difficulty is to characterize the bias in a tractable way under such algorithms, where the adaptive data collection procedure induces complex dependency among rewards. 
Another interesting direction is to consider the finite-sample analysis of the bias reduction via bootstrap,
which is more relevant given that the bias is typically larger in small samples.
This may require a different set of mathematical techniques, and we leave it for future work. 


\newpage
\medskip
\bibliographystyle{abbrv}
\bibliography{ref}

\newpage
\appendix

\section{Additional experiments} \label{sec:four-armed}
We also conduct additional bandit experiments with $K=4$ arms:
Gaussian reward with mean $\mu=\{2,2.5,3,3.5\}$ and standard deviation $\sigma=\{2,1,2,1\}$, and Bernoulli reward with mean $\mu=\{0.4, 0.5, 0.7, 0.8\}$.
The four bandit algorithms (ETC, UCB, TS, EG) with the same parameters setting as Section \ref{sec:numerical} are repeated $1,000$ times, each instance with $B=1,000$ bootstrap simulations for both Efron's bootstrap and Gaussian multiplier bootstrap.
The results are presented below.
They further demonstrate the effectiveness of the proposed algorithm.
\begin{table}[h!]
\centering
\small{
\begin{tabular}{ccccccccc}
    \toprule
    \multirow{3}{*}{Algorithm} & \multirow{3}{*}{Reward} & \multirow{3}{*}{Arm}  &\multicolumn{2}{c}{True Environment} &\multicolumn{2}{c}{MB Environment} &\multicolumn{2}{c}{EB Environment}\\
	&&&\multirow{2}{*}{$\E[\hat{\mu}_k]-\mu_k$} &\multirow{2}{*}{$\E[\hat{\mu}_k]$} &Estimated &Corrected  &Estimated &Corrected \\
	&&&&&Bias&$\E[\hat{\mu}_k]$&Bias&$\E[\hat{\mu}_k]$\\\hline
	\multirow{8}{*}{ETC} &\multirow{4}{*}{Normal} &Arm1 &-0.0264 &1.9736 &-0.0398 &2.0134 &-0.0385 &2.0121\\
	&&Arm2 &-0.0071 &2.4929 &-0.0157 &2.5087 &-0.0157 &2.5087\\
	&&Arm3 &-0.0416 &2.9584 &-0.0375 &2.9959 &-0.0369 &2.9952\\
	&&Arm4 &-0.1924 &3.3076 &-0.1443 &3.4519 &-0.1448 &3.4524\\\cline{2-9}
	&\multirow{4}{*}{Bernoulli} &Arm1 &-0.006 &0.394 &-0.0064 &0.4005 &-0.0092 &0.4032\\
	&&Arm2 &-0.0102 &0.4898 &-0.0127 &0.5024 &-0.0158 &0.5055\\
	&&Arm3 &-0.0366 &0.6634 &-0.0263 &0.6897 &-0.0284 &0.6918\\
	&&Arm4 &-0.029 &0.771 &-0.0259 &0.797 &-0.0273 &0.7983\\
	\bottomrule
	\multirow{8}{*}{UCB} &\multirow{4}{*}{Normal} &Arm1 &-0.8538 &1.1462 &-0.2737 &1.4198 &-0.3998 &1.546\\
	&&Arm2 &-0.7435 &1.7565 &-0.2109 &1.9674 &-0.2136 &1.9701\\
	&&Arm3 &-0.5521 &2.4479 &-0.1479 &2.5959 &-0.1482 &2.5961\\
	&&Arm4 &-0.699 &2.801 &-0.4529 &3.2539 &-0.5017 &3.3026\\\cline{2-9}
	&\multirow{4}{*}{Bernoulli} &Arm1 &-0.0661 &0.3339 &-0.0564 &0.3904 &-0.0503 &0.3843\\
	&&Arm2 &-0.0666 &0.4334 &-0.0580 &0.4914 &-0.0539 &0.4873\\
	&&Arm3 &-0.0393 &0.6607 &-0.0363 &0.6969 &-0.0380 &0.6986\\
	&&Arm4 &-0.0119 &0.7881 &-0.0121 &0.8002 &-0.0137 &0.8018\\
	\bottomrule
	\multirow{8}{*}{TS} &\multirow{4}{*}{Normal} &Arm1 &-0.7917 &1.2083 &-0.2996 &1.5079 &-0.3321 &1.5404\\
	&&Arm2 &-0.2872 &2.2128 &-0.1701 &2.3829 &-0.1721 &2.3848\\
	&&Arm3 &-0.2594 &2.7406 &-0.1401 &2.8807 &-0.1398 &2.8804\\
	&&Arm4 &-0.6559 &2.8441 &-0.4913 &3.3354 &-0.4974 &3.3415\\\cline{2-9}
	&\multirow{4}{*}{Bernoulli} &Arm1 &-0.1112 &0.2888 &-0.0855 &0.3743 &-0.0710 &0.3598\\
	&&Arm2 &-0.0924 &0.4076 &-0.0670  &0.4746 &-0.6560 &0.4732\\
	&&Arm3 &-0.0748 &0.6252 &-0.0643 &0.6895 &-0.0674 &0.6926\\
	&&Arm4 &-0.0278 &0.7722 &-0.0266 &0.7987 &-0.0315 &0.8037\\
	\bottomrule
	\multirow{8}{*}{EG} &\multirow{4}{*}{Normal} &Arm1 &-0.2252 &1.7748 &-0.2128 &1.9876 &-0.2204 &1.9952\\
	&&Arm2 &-0.1482 &2.3518 &-0.1039 &2.4556 &-0.1078 &2.4595\\
	&&Arm3 &-0.3591 &2.6409 &-0.2167 &2.8576 &-0.2235 &2.8644\\
	&&Arm4 &-0.959 &2.541 &-0.5405 &3.0815 &-0.5554 &3.0964\\\cline{2-9}
	&\multirow{4}{*}{Bernoulli} &Arm1 &-0.1115 &0.2885 &-0.0652 &0.3537 &-0.0704 &0.3590\\
	&&Arm2 &-0.1310 &0.3690 &-0.0820 &0.4510 &-0.0957 &0.4646\\
	&&Arm3 &-0.1517 &0.5483 &-0.0985 &0.6467 &-0.1168 &0.6650\\
	&&Arm4 &-0.1182 &0.6818 &-0.0831 &0.7649 &-0.1083 &0.7901\\
	\bottomrule
    \end{tabular}
    }
    \vspace{2mm}
    \caption{Debiasing using Bootstrap for a Four-Armed Bandit Problem.}
    \label{table:four_arm}
\end{table}

\section{Proof of Lemma \ref{lem:bias_negative}} \label{sec:lemma1}
\begin{proof}
For arm $k$, the average rewards of the whole decision horizon $T$ is 
\begin{align}
\hat \mu_{k}=
\left\{ \begin{array}{ll}
\frac{\sum_{t=(k-1)m+1}^{km} r_t + \sum_{t=Km+1}^{T} r_t} {m + T - m K}, &\quad \text{if $k=a_{mK+1}$,}\\
 \frac{\sum_{t=(k-1)m+1}^{km} r_t} {m} = \hat \mu_{k}(mK), &\quad \text{if $k \ne a_{mK+1}$}.
\end{array} \right.
\end{align}
Hence we have
\begin{align}
\E_P[\hat \mu_{k}] = \E_P \left[  \hat \mu_{k}(mK) \I{k \ne a_{mK+1}} +  \hat \mu_{k}  \I{k = a_{mK+1}} \right].
\end{align}
Note for each $k=1, \ldots, K$, $\E_P[\hat \mu_{k}(mK)] = \mu_k$. Then the bias of arm $k$ is given by
\begin{align} \label{eq:bias-k}
\E_P[\hat \mu_{k}] - \mu_k &= \E_P \left[ (- \hat \mu_{k}(mK) +   \hat \mu_{k} ) \I{k = a_{mK+1}} \right] \nonumber \\
&=  \frac{T-mK}{m + T - m K} \cdot \E_P \left[ \left(  \frac{\sum_{t=Km+1}^{T} r_t}{T-mK}  - \hat \mu_{k}(mK) \right) \I{k = a_{mK+1}} \right]  \nonumber \\
& =   \frac{T-mK}{m + T - m K} \cdot \E_P \left[ \left( \mu_k  - \hat \mu_{k}(mK) \right) \I{k = a_{mK+1}} \right],
\end{align}
where the expectation is completely determined by the joint distribution of $(\hat \mu_{k}(mK))_i$ with independent marginals. In particular, the marginal distribution of $\hat \mu_{k}(mK)$ is simply the average of $m$ i.i.d. rewards associated with arm $k$. Mathematically, we have
\begin{align*}
\E_P[\hat \mu_{k}] - \mu_k
&= \frac{T-mK}{m + T - m K} \cdot \int_{\mathbb{R}} (\mu_k - x) \mathbb{P} (k = a_{mK+1})  \mathbb{P} (\hat \mu_{k}(mK) \in dx )\\
&\leq \frac{T-mK}{m + T - m K} \cdot \int_{\mathbb{R}} (\mu_k - x) \mathbb{P} (\max_{k' \ne k} \hat \mu_{k'}(mK) \leq x )  \mathbb{P} (\hat \mu_{k}(mK) \in dx ).
\end{align*}
Therefore, the bias is negative since
$\int_{\mathbb{R}} (\mu_k - x)   \mathbb{P} (\hat \mu_{k}(mK) \in dx ) = \mu_k - \E_P[\hat \mu_{k}(mK)] = 0$, and
 $\mathbb{P} (\max_{k' \ne k} \hat \mu_{k'}(mK) \leq x )$ puts more weights on larger values of $x$.
\end{proof}

\section{Proof of Proposition \ref{prop:two_normal_true_bias}}
\begin{proof}
Recall for a general $K-$armed bandit problems with continuous rewards, the bias of the sample mean of arm $k$ has the following expression:
\begin{align*}
\E_P[\hat \mu_{k}] - \mu_k =   \frac{T-mK}{m + T - m K} \cdot \E_P \left[ ( \mu_k  - \hat \mu_{k}(mK) ) \I{ \hat \mu_{k}(mK) > \max_{k' \ne k} \hat \mu_{k'}(mK)} \right].
\end{align*}
To prove Proposition~\ref{prop:two_normal_true_bias} with $K=2$, we first consider the special case where the arm 1 follows $N(\mu_1, \sigma_1^2)$ while the arm 2 has deterministic rewards $\mu_2,$ i.e. $\sigma_2=0.$
Then we have the bias of arm 2 is clearly zero, and the bias of arm 1 is given by
\begin{align}\label{arm1bias}
\E_P[\hat \mu_{1}] - \mu_1 =   \frac{T-2m}{T - m} \cdot \E_P \left[ ( \mu_1  - \hat{\mu}_1(2m)) \I{ \hat{\mu}_1(2m) \ge \mu_2}\right],
\end{align}
where $\hat{\mu}_1(2m) = \frac{1}{m} \sum_{i=1}^m {r_i}$ with $r_i$ generated from $N(\mu_1, \sigma_1^2)$.
Note that $\hat{\mu}_1(2m)$ is a random variable following the normal distribution $N(\mu_1, \frac{\sigma_1^2}{m})$.
Hence we obtain
\begin{align}\label{arm1bias:arm2deterministic}
 \E_P \left[ ( \mu_1  - \hat{\mu}_1(2m)) \I{ \hat{\mu}_1(2m) \ge \mu_2}\right] = \frac{\sigma_1}{\sqrt{m}}\cdot \E \left[ -Z \cdot \I{ Z > \frac{\mu_2- \mu_1}{\sigma_1/ \sqrt{m}} } \right],
\end{align}
where $Z \sim N(0,1).$ Notice that $\E \left[ Z \cdot \I{Z > c} \right] = \phi(c)$ for any $c$,
where $\phi$ is the density function of $N(0,1).$
It follows that
\begin{align}
 \E_P \left[ ( \mu_1  - \hat{\mu}_1(2m)) \I{ \hat{\mu}_1(2m) \ge \mu_2}\right]= -  \frac{\sigma_1}{\sqrt{m}} \phi \left( \frac{\mu_2- \mu_1}{ \sigma_1/ \sqrt{m}} \right)=-  \frac{\sigma_1}{\sqrt{ 2 \pi m}} \exp \left[ - \frac{1}{2} \left( \frac{\mu_2- \mu_1}{ \sigma_1} \right)^2 m \right].
\end{align}
On combining with \eqref{arm1bias}, we immediately obtain the expression for the bias of arm 1 in \eqref{eq:bias-2arm-normal}.

Next we consider the setting where the reward of arm 2 follows $N(\mu_2, \sigma_2^2)$ with $\sigma_2>0$. The bias of arm 1 is clearly given by
\begin{align}\label{arm1bias-sigma}
\E_P[\hat \mu_{1}] - \mu_1 =   \frac{T-2m}{ T - m } \cdot \E_P \left[ ( \mu_1  - \hat{\mu}_1(2m)) \I{ \hat{\mu}_1(2m) > \hat{\mu}_2(2m)} \right],
\end{align}
where $\hat{\mu}_2(2m) \sim N(\mu_2, \frac{\sigma_2^2}{m})$ since $\hat{\mu}_1(2m) = \frac{1}{m} \sum_{i=m+1}^{2m} r_i$ with $r_i$ obtained from $N(\mu_2, \sigma_2^2)$.
The difference between \eqref{arm1bias} and \eqref{arm1bias-sigma} is just that we substitute a deterministic value $\mu_2$ by a normal random variable $\hat{\mu}_2(2m)$. So it follows that
\begin{align}
&\E_P \left[ ( \mu_1  - \hat{\mu}_1(2m)) \I{\hat{\mu}_1(2m)> \hat{\mu}_2(2m)} \right]\\
& = \int_y \left[ -  \frac{\sigma_1}{\sqrt{ 2 \pi m}} \exp \left[ - \frac{1}{2} \left( \frac{y- \mu_1}{ \sigma_1} \right)^2 m \right] \cdot \phi\left(  \frac{ y - \mu_2}{ \sigma_2/ \sqrt{m}} \right) \cdot \frac{\sqrt{m}}{\sigma_2}\right] dy\\
&=-  \frac{\sigma_1}{ 2\pi \sigma_2}  \int_y
\exp \left[ - \frac{m}{2 \sigma_1^2} (y- \mu_1)^2 -\frac{m}{2\sigma_2^2} (y - \mu_2)^2 \right] dy.
\end{align}
Note that
\begin{align*}
- \frac{m}{2 \sigma_1^2} (y- \mu_1)^2 -\frac{m}{2\sigma_2^2} (y - \mu_2)^2 = -\frac{m}{2\sigma_1^2 \sigma_2^2}\left[ \left(\sqrt{\sigma_1^2+\sigma_2^2}y - \frac{\sigma_2^2 \mu_1 +\sigma_1^2 \mu_2}{\sqrt{\sigma_2^2+\sigma_1^2}}\right)^2+\frac{\sigma_2^2\sigma_1^2 }{\sigma_2^2+\sigma_1^2}(\mu_1-\mu_2)^2\right].
\end{align*}
Hence, we can obtain
\begin{align*}
&\E_P \left[ ( \mu_1  - \hat{\mu}_1(2m)) \I{\hat{\mu}_1(2m) > \hat{\mu}_2(2m)} \right]\\
&=-\frac{\sigma_1}{ 2 \pi\sigma_2}  \int_y \exp \left[ -\frac{m}{2\sigma_1^2 \sigma_2^2}\left(\sqrt{\sigma_1^2+\sigma_2^2}y - \frac{\sigma_2^2 \mu_1 +\sigma_1^2 \mu_2}{\sqrt{\sigma_2^2+\sigma_1^2}}\right)^2 \right] dy \cdot \exp \left[ -\frac{m}{2(a^2+\sigma^2)} (\mu_1-\theta)^2\right]\\
&=-\frac{\sigma_1^2}{\sqrt{2\pi(\sigma_1^2+\sigma_2^2)m}}\exp \left[ -\frac{m}{2(\sigma_1^2+\sigma_2^2)} (\mu_1-\mu_2)^2\right].
\end{align*}
Therefore, the bias of arm 1 is
\begin{align}
    \E_P[\hat \mu_1] - \mu_1=-\frac{T-2m}{ T - m}\frac{\sigma
    _1^2}{\sqrt{2\pi(\sigma_1^2+\sigma_2^2)m}}\exp \left[ -\frac{m}{2(\sigma_1^2+\sigma_2^2)} (\mu_1-\mu_2)^2\right].
\end{align}
This proves \eqref{eq:bias-2arm-normal} for $k=1$.

Similarly, one can compute the bias of arm 2 by
\begin{align*}
\E_P[\hat \mu_{2}] - \mu_2 =   \frac{T-2m }{ T - m } \cdot \E \left[ ( \mu_2  - \hat{\mu}_2(2m)) \I{\hat{\mu}_2(2m) > \hat{\mu}_1(2m) } | P \right],
\end{align*}
where $\hat{\mu}_1(2m) = \frac{1}{m} \sum_{i=1}^m {r_i} \sim N(\mu_1, \frac{\sigma_1^2}{m})$ and $\hat{\mu}_2(2m) = \frac{1}{m} \sum_{i=m+1}^{2m} r_i \sim N(\mu_2, \frac{\sigma_2^2}{m})$. Using a similar argument, we can obtain that the bias of arm 2 is
\begin{align}
    \E_P[\hat \mu_2] - \mu_2=-\frac{T-2m}{ T - m }\frac{\sigma
    _2^2}{\sqrt{2\pi(\sigma_1^2+\sigma_2^2)m}}\exp \left[ -\frac{m}{2(\sigma_1^2+\sigma_2^2)} (\mu_1-\mu_2)^2\right].
\end{align}
Therefore, the proof is complete.
\end{proof}

\section{Proof of Theorem \ref{thm:decayrate_normal}} \label{sec:thm1}
\begin{proof}
We prove the result for $k=1$ as the proof for $k=2$ is similar.
Recall
\begin{align} \label{eq:g-1}
    g_1(\mu_1, \mu_2, \sigma_1^2, \sigma_2^2) = \log \left( \frac{T-2 m}{ T - m }\frac{\sigma
    _1^2}{\sqrt{2\pi(\sigma_1^2+\sigma_2^2)m}}\exp \left[ -\frac{m}{2(\sigma_1^2+\sigma_2^2)} (\mu_1-\mu_2)^2\right] \right).
\end{align}
To prove the result, we will perform Taylor expansion of the (random) function $g_1(\hat{\mu}_1, \hat{\mu}_2, \hat{\sigma}_1^2, \hat{\sigma}_2^2)$ at the point $(\mu_1, \mu_2, \sigma_1^2, \sigma_2^2)$. We can readily compute that the first order (partial) derivatives of $g_1$ in \eqref{eq:g-1} are given by
\begin{align*}
&\frac{\partial g_1}{\partial \mu_1}(\mu_1, \mu_2,\sigma_1^2, \sigma_2^2)= -\frac{m(\mu_1-\mu_2)}{\sigma_1^2+\sigma_2^2},\\
&\frac{\partial g_1}{\partial \mu_2}(\mu_1, \mu_2, \sigma_1^2, \sigma_2^2)= \frac{m(\mu_1-\mu_2)}{\sigma_1^2+\sigma_2^2},\\
&\frac{\partial g_1}{\partial \sigma_1^2}(\mu_1, \mu_2, \sigma_1^2, \sigma_2^2)=\frac{1}{\sigma_1^2}-\frac{1}{2(\sigma_1^2+\sigma_2^2)}+\frac{ m(\mu_1-\mu_2)^2}{2(\sigma_1^2+\sigma_2^2)^2},\\
&\frac{\partial g_1}{\partial \sigma_2^2}(\mu_1, \mu_2, \sigma_1^2, \sigma_2^2)=-\frac{1}{2(\sigma_1^2+\sigma_2^2)}+\frac{ m(\mu_1-\mu_2)^2}{2(\sigma_1^2+\sigma_2^2)^2}.
\end{align*}
Hence we can write $\nabla g_1(\mu_1,\mu_2,\sigma_1^2,\sigma_2^2)=\left(
\begin{matrix}
-d_1m  \\
d_1m\\
d_2m+e_1 \\
d_2m-e_2
\end{matrix}
\right)$,
where $d_1, d_2, e_1, e_2$ are nonzero constants that are independent of $m$.

To perform the Taylor expansion, we recall and introduce the stochastic ``o'' symbol, see \cite{van2000asymptotic}.
For a sequence of random variables $\{X_n, n=0,1,2,\ldots\}$, the expression $X_n=o_P(1)$ denotes that $(X_n)$ converges to zero in probability.
Furthermore, for a given sequence of random variables $\{R_n, n=0,1,2,\ldots\}$, the notation $X_n=o_P(R_n)$ means $X_n=Y_nR_n$ with $Y_n=o_P(1)$. We will show later that $||(\hat{\mu}_1-\mu_1,
\hat{\mu}_2-\mu_2,
\hat{\sigma}_1^2-\sigma_1^2,
\hat{\sigma}_2^2-\sigma_2^2)|| =o_P(1)$ as $m \rightarrow \infty$.
Then Lemma 2.12 of \cite{van2000asymptotic} guarantees that we can take the Taylor expansion of $g_1(\hat{\mu}_1, \hat{\mu}_2, \hat{\sigma}_1^2, \hat{\sigma}_2^2)$ at point $(\mu_1, \mu_2, \sigma_1^2, \sigma_2^2)$:
\begin{align*}
g_1(\hat{\mu}_1, \hat{\mu}_2, \hat{\sigma}_1^2, \hat{\sigma}_2^2)=g_1(\mu_1, \mu_2, \sigma_1^2, \sigma_2^2)+\nabla g_1(\mu_1,\mu_2,\sigma_1^2,\sigma_2^2)^T
\left(
\begin{matrix}
\hat{\mu}_1-\mu_1  \\
\hat{\mu}_2-\mu_2 \\
\hat{\sigma}_1^2-\sigma_1^2 \\
\hat{\sigma}_2^2-\sigma_2^2
\end{matrix}
\right)
+o_P\left( \left\|
\begin{matrix}
\hat{\mu}_1-\mu_1  \\
\hat{\mu}_2-\mu_2 \\
\hat{\sigma}_1^2-\sigma_1^2 \\
\hat{\sigma}_2^2-\sigma_2^2
\end{matrix}\right\|
\right).
\end{align*}
It follows that
\begin{align}
\frac{g_1(\hat{\mu}_1, \hat{\mu}_2, \hat{\sigma}_1^2, \hat{\sigma}_2^2)}{g_1(\mu_1, \mu_2, \sigma_1^2, \sigma_2^2)}=1+\frac{\nabla g_1(\mu_1,\mu_2,\sigma_1^2,\sigma_2^2)^T
\left(
\begin{matrix}
\hat{\mu}_1-\mu_1  \\
\hat{\mu}_2-\mu_2 \\
\hat{\sigma}_1^2-\sigma_1^2 \\
\hat{\sigma}_2^2-\sigma_2^2
\end{matrix}
\right)}{g_1(\mu_1, \mu_2, \sigma_1^2, \sigma_2^2)} + \frac{o_P\left( \left\|
\begin{matrix}
\hat{\mu}_1-\mu_1  \\
\hat{\mu}_2-\mu_2 \\
\hat{\sigma}_1^2-\sigma_1^2 \\
\hat{\sigma}_2^2-\sigma_2^2
\end{matrix}\right\|
\right)}{g_1(\mu_1, \mu_2, \sigma_1^2, \sigma_2^2)}.\label{equ:taylor}
\end{align}

To prove Theorem~\ref{thm:decayrate_normal}, we proceed to show the second and the third terms of \eqref{equ:taylor} will both converge to zero in probability as $m \rightarrow \infty$. For notational convenience,
in the following we use $\hat{g}_1, g_1$ and $\nabla g_1$ to denote $g_1(\hat{\mu}_1, \hat{\mu}_2, \hat{\sigma}_1^2, \hat{\sigma}_2^2), g_1(\mu_1, \mu_2, \sigma_1^2, \sigma_2^2)$ and $\nabla g_1(\mu_1,\mu_2,\sigma_1^2,\sigma_2^2)$, respectively.

We first show the second term of \eqref{equ:taylor} converges to 0 in probability. Note that
\begin{align}\label{inequ:CS}
\left|\frac{1}{g_1} \cdot {\nabla g_1^T
\left(
\begin{matrix}
\hat{\mu}_1-\mu_1  \\
\hat{\mu}_2-\mu_2 \\
\hat{\sigma}_1^2-\sigma_1^2 \\
\hat{\sigma}_2^2-\sigma_2^2
\end{matrix}
\right)} \right|
\leq
\left\|
\begin{matrix}
-d_1m/g_1  \\
d_1m/g_1\\
(d_2m+e_1)/g_1 \\
(d_2m-e_2)/g_1
\end{matrix}
\right\|
\cdot
\left\|
\begin{matrix}
\hat{\mu}_1-\mu_1  \\
\hat{\mu}_2-\mu_2 \\
\hat{\sigma}_1^2-\sigma_1^2 \\
\hat{\sigma}_2^2-\sigma_2^2
\end{matrix}\right\|.
\end{align}
It is easy to see from \eqref{eq:g-1} that $g_1=\Theta(m)$ as $m \rightarrow \infty$. Then there exists a constant $0<C<\infty$ such that
$\lim_{m \to \infty}
\left\|
\begin{matrix}
-d_1m/g_1 \\
d_1m/g_1\\
(d_2m+e_1)/g_1\\
(d_2m-e_2)/g_1
\end{matrix}
\right\| \le C$. Hence it suffices to show
\begin{align}\label{Pconverge:diff}
\lim \limits_{m \to \infty} \mathbb{P} \left(\left\|
\begin{matrix}
\hat{\mu}_1-\mu_1  \\
\hat{\mu}_2-\mu_2 \\
\hat{\sigma}_1^2-\sigma_1^2 \\
\hat{\sigma}_2^2-\sigma_2^2
\end{matrix}\right\| \geq \epsilon \right) \to 0, \quad \emph{i.e.,} \quad
\left\|
\begin{matrix}
\hat{\mu}_1-\mu_1  \\
\hat{\mu}_2-\mu_2 \\
\hat{\sigma}_1^2-\sigma_1^2 \\
\hat{\sigma}_2^2-\sigma_2^2
\end{matrix}\right\|
\stackrel{P}{\longrightarrow} 0.
\end{align}
To this end, we can compute that
\begin{align*}
&\mathbb{P} \left(\left\|
\begin{matrix}
\hat{\mu}_1-\mu_1  \\
\hat{\mu}_2-\mu_2 \\
\hat{\sigma}_1^2-\sigma_1^2 \\
\hat{\sigma}_2^2-\sigma_2^2
\end{matrix}\right\| \geq \epsilon \right) \\
&\leq \mathbb{P}\left( |\hat{\mu}_1-\mu_1|+|\hat{\mu}_2-\mu_2|+ |\hat{\sigma}_1^2-\sigma_1^2| + |\hat{\sigma}_2^2-\sigma_2^2| \geq \epsilon \right)\\
&\leq \mathbb{P}\left( |\hat{\mu}_1-\mu_1| \geq \frac{\epsilon}{4}\right)+\mathbb{P}\left(|\hat{\mu}_2-\mu_2|\geq \frac{\epsilon}{4}\right)+ \mathbb{P}\left( |\hat{\sigma}_1^2-\sigma_1^2|\geq \frac{\epsilon}{4}\right) + \mathbb{P} \left(|\hat{\sigma}_2^2-\sigma_2^2| \geq \frac{\epsilon}{4} \right).
\end{align*}
We know that
\begin{align*}
    &\mathbb{P}\left( |\hat{\mu}_1-\mu_1| \geq \frac{\epsilon}{4}\right)\\
    &=\mathbb{P}\left( |\hat{\mu}_1-\mu_1| \geq \frac{\epsilon}{4}\right|\text{arm 1 chosen})\mathbb{P}(\text{arm 1 chosen})+\mathbb{P}\left( |\hat{\mu}_1-\mu_1| \geq \frac{\epsilon}{4}\right|\text{arm 2 chosen})\mathbb{P}(\text{arm 2 chosen})\\
    &\leq \mathbb{P}\left( \left|\frac{x_1+\ldots+x_{T-m}}{T-m}-\mu_1\right| \geq \frac{\epsilon}{4}\right)+\mathbb{P}\left( \left|\frac{x_1+\ldots+x_{m}}{m}-\mu_1\right| \geq \frac{\epsilon}{4}\right)\\
    &\leq \frac{16\sigma_1^2}{(T-m)\epsilon^2}+\frac{16\sigma_1^2}{m\epsilon^2},
\end{align*}
where the last inequality comes from Chebyshev's inequality.
Therefore, we can obtain $\lim \limits_{m \to \infty} \mathbb{P}\left( |\hat{\mu}_1-\mu_1| \geq \frac{\epsilon}{4}\right) \to 0$, since $T - m \ge m$. Similarly, we can obtain that $\lim \limits_{m \to \infty} \mathbb{P}\left( |\hat{\mu}_2-\mu_2| \geq \frac{\epsilon}{4}\right) \to 0, \lim \limits_{m \to \infty} \mathbb{P}\left( |\hat{\sigma}_1^2-\sigma_1^2| \geq \frac{\epsilon}{4}\right) \to 0$ and $\lim \limits_{m \to \infty} \mathbb{P}\left( |\hat{\sigma}_2^2-\sigma_2^2| \geq \frac{\epsilon}{4}\right) \to 0$. This establishes \eqref{Pconverge:diff}, and it follows that the second term of \eqref{equ:taylor} converges to 0 in probability.

Finally, we show that the third term of \eqref{equ:taylor} also converges to 0 in probability. From \eqref{Pconverge:diff}, we have $o_P\left( \left\|
\begin{matrix}
\hat{\mu}_1-\mu_1  \\
\hat{\mu}_2-\mu_2 \\
\hat{\sigma}_1^2-\sigma_1^2 \\
\hat{\sigma}_2^2-\sigma_2^2
\end{matrix}\right\|
\right)\stackrel{P}{\longrightarrow} 0$. Moreover, $g_1=\Theta(m)$ which implies $\lim \limits_{m \to \infty} \frac{1}{g_1}=0$, then it follows that
\begin{align}
\frac{o_P\left( \left\|
	\begin{matrix}
	\hat{\mu}_1-\mu_1  \\
	\hat{\mu}_2-\mu_2 \\
	\hat{\sigma}_1^2-\sigma_1^2 \\
	\hat{\sigma}_2^2-\sigma_2^2
	\end{matrix}\right\|
	\right)}{g_1} \stackrel{P}{\longrightarrow} 0.
\end{align}
Therefore, we can deduce from \eqref{equ:taylor} that the result in Theorem~\ref{thm:decayrate_normal} holds.
\end{proof}

\section{Proof of Proposition \ref{prop:two_subGaussian_true_bias}} \label{sec:subgaussian}
\begin{proof}
Recall from \eqref{eq:bias-k} that the bias of arm 1 is given by
\begin{align}\label{eq:bias1-G}
\E_P[\hat \mu_{1}(T)] - \mu_1 =   \frac{T-2m}{T-m} \cdot \E_P \left[ ( \mu_1  - \bar X_m) \I{ \bar X_m \ge \mu_2}\right].
\end{align}
We focus on estimating the quantity
\begin{align}
G( \bar X_m) \coloneqq \E_P \left[ ( \mu_1  - \bar X_m) \I{ \bar X_m \ge \mu_2 } \right].
\end{align}
We first consider the case $\mu_1 < \mu_2$. Then
\begin{align}\label{arm1bias:generalreward}
    G( \bar X_m)=\E \left[ ( \mu_2  - \bar X_m) \I{ \bar X_m \ge \mu_2 }  \right]  +  (\mu_1 - \mu_2) \mathbb{P} (\bar X_m \ge \mu_2).
\end{align}

Since $\mu_1 < \mu_2$, for the second term of \eqref{arm1bias:generalreward}, by the Bahadur-Rao theorem (Theorem 3.7.4 \cite{dembo1998large}), we have
\begin{align}\label{BR:tailpro}
 \mathbb{P} (\bar X_m \ge \mu_2) = \frac{e^{-m \Lambda^*(\mu_2)}}{ \sqrt{2 \pi m \eta''\left(\zeta  \right)}} (c_0 + o(1)),
\end{align}
where $\eta (h) = \log  \E [e^{h X_1}]$, $\zeta>0$ is defined by the implicit equation $\eta'\left(\zeta  \right) =\mu_2$ for a given $\mu_2> \mu_1$, and $\Lambda^*$ is the Legendre-Fenchel transform of $\eta$:
\begin{align*}
\Lambda^*(x) = \sup_{h \in \mathbb{R}} ( hx - \eta(h) ).
\end{align*}
The constant $c_0>0$ depends on the reward distribution
$X_i$ is lattice or non-lattice, and it is given by
\begin{align} \label{eq:c0}
c_0=
\left\{ \begin{array}{ll}
\frac{1}{|\zeta|}, &\text{if $X_i$ has a non-lattice distribution,}\\
\multirow{2}{*}{$\frac{d}{1-e^{-|\zeta|  d}}$,} &\text{if $X_i$ has a lattice distribution, so that $\mathbb{P}(X_i = \mu_2) \in (0,1)$, and $d$ is the largest number to}\\
&\text{make $(X_i-\mu_2)/d$ is (a.s.) an integer number}.
\end{array} \right.
\end{align}

In addition, for the first term of \eqref{arm1bias:generalreward}, we also have
the following result, the proof of which is deferred to the end of this section.
\begin{lemma}\label{Lemma:BR_extension1}
Let $X_i$ be i.i.d. random variables with mean $\mu_1$ and logarithmic moment generating function $\eta (h) = \log  \E [e^{h X_i}]$ and $\zeta $ is defined by the implicit equation $\eta'\left(\zeta  \right) =\mu_2$ for a given $\mu_2>\mu_1$. Let $J_m(\mu_2)=-\zeta^2 \sqrt{2\pi m \eta''\left(\zeta  \right)}m \exp \left(m \Lambda^*(\mu_2)\right)$.
\begin{itemize}
    \item[(a)]If the law of $X_i$ is non-lattice, then
    \begin{align}
    \lim_{m \to \infty} J_m(\mu_2)\E_P [(\mu_2-\bar X_m) \1_{\{\bar X_m \ge \mu_2\}}]=1.
    \end{align}
    \item[(b)] If $X_i$ has a lattice law, so that $\mathbb{P}(X_i = \mu_2) \in (0,1)$,  $(X_i-\mu_2)/d$ is (a.s.) an integer number, and $d$ is the largest number with this property, then
    \begin{align}
    \lim_{m \to \infty} J_m(\mu_2)\E_P [(\mu_2-\bar X_m) \1_{\{\bar X_m \ge \mu_2\}}]=\frac{\zeta d \cdot e^{-\zeta d}}{1-e^{-\zeta d}}.
    \end{align}
\end{itemize}
\end{lemma}

Therefore, we have
\begin{align*}
\E_P \left[ ( \mu_2  - \bar X_m) \I{ \bar X_m \ge \mu_2 }  \right] &= \frac{e^{-m \Lambda^*(\mu_2)}}{ \sqrt{2 \pi m^3 \eta''(\zeta )}} (c_1 + o(1)),
\end{align*}
where $c_1$ is an explicit constant which also depends on whether the reward $X_i$ is lattice or non-lattice:
\begin{align}
c_1=
\left\{ \begin{array}{ll}
-\frac{1}{\zeta^2}, &\text{if $X_i$ has a non-lattice distribution,}\\
\multirow{2}{*}{$-\frac{d \cdot e^{-\zeta  d}}{(1-e^{-\zeta  d})\zeta}$,} &\text{if $X_i$ has a lattice distribution, so that $\mathbb{P}(X_i = \mu_2) \in (0,1)$, and $d$ is the largest}\\
&\text{number to make $(X_i-\mu_2)/d$ is (a.s.) an integer number}.
\end{array} \right.\\ \label{eq:c1}
\end{align}

Hence we obtain when $\mu_1 < \mu_2,$
\begin{align} \label{eq:case1-G}
G( \bar X_m) =  \frac{e^{-m \Lambda^*(\mu_2)}}{ \sqrt{2 \pi m \eta''\left(\zeta  \right)}} \left(c_0(\mu_1 - \mu_2)+\frac{c_1}{m}+o(1) \right) = \frac{e^{-m \Lambda^*(\mu_2)}}{ \sqrt{2 \pi m \eta''\left(\zeta  \right)}} \left(c_0(\mu_1 - \mu_2)+o(1) \right),
\end{align}
where $c_0>0$ is given in \eqref{eq:c0}. Hence we have completed the proof for the case $\mu_1 < \mu_2$.

We next consider the case when $\mu_1 > \mu_2$. Then $\{\bar X_m \ge \mu_2\}$ is not a rare event anymore when $m$ is large, but  $\{\bar X_m \le \mu_2\}$ is a rare event with small probabilities. We can compute
\begin{align}\label{eq:case2}
G( \bar X_m) &=\E_P \left[ ( \mu_1  - \bar X_m) \I{ \bar X_m \ge \mu_2 } \right] \nonumber \\
&= 0 - \E_P \left[ ( \mu_1  - \bar X_m) \I{ \bar X_m \le \mu_2 } \right] \nonumber \\
& = (\mu_2 - \mu_1)  \mathbb{P} (\bar X_m \le \mu_2) + \E_P \left[ (\bar X_m-\mu_2) \I{ \bar X_m \le \mu_2 } \right],
\end{align} 
where the second equality is due to the fact that $\E_P( \mu_1  - \bar X_m)=0.$ Applying Bahadur-Rao theorem to compute the first term in \eqref{eq:case2} and use a similar argument as in Lemma~\ref{Lemma:BR_extension1} to compute the second term in \eqref{eq:case2}, we can similarly obtain that when $\mu_1 > \mu_2,$
\begin{align} \label{eq:case2-G}
G( \bar X_m) =   \frac{e^{-m
 \Lambda^*(\mu_2)}}{ \sqrt{2 \pi m \eta''(\zeta)}}  (c_0 \cdot (\mu_2 - \mu_1) + o(1)),
\end{align}
where $\zeta$ is the solution to the equation $\eta'(\zeta) =\mu_2$ and now $\zeta<0$ when $\mu_1 > \mu_2.$ The constant $c_0>0$ is given in \eqref{eq:c0}. 

Finally, we can combine \eqref{eq:bias1-G}, \eqref{eq:case1-G} and \eqref{eq:case2-G} to infer that the bias of arm 1 is given by
\begin{align*}
\E_P[\hat \mu_{1}] - \mu_1 =   \frac{T-2m}{T - m } \frac{e^{-m \Lambda^*(\mu_2)}}{ \sqrt{2 \pi m \eta''(\zeta )}}  (-c_* + o(1)),
\end{align*}
as $m \rightarrow \infty$, where the constant $c_*$ is given by
\begin{align} \label{eq:c_*}
    c_* = c_0 \cdot |\mu_1 -\mu_2| >0, 
\end{align}
with $c_0>0$ given in \eqref{eq:c0}.
The proof is therefore complete.
\end{proof}

\subsection{Proof of Lemma \ref{Lemma:BR_extension1}}
\begin{proof}
We adapt the proof of Bahadur-Rao theorem (Theorem 3.7.4 \cite{dembo1998large}) to our setting. When $\mu_1<\mu_2$, recall $\eta (h) = \log  \E [e^{h X_1}]$, and $\Lambda^*(x)=\sup_{h \in \mathbb{R}} ( hx - \eta(h) )$. Then define $D_\eta \coloneqq \{h \in \R : \eta(h) < \infty\}$. Then in $\text{Int}\{D_{\eta}\}$, we have $\eta'(h)=\E[X_1 \exp (hX_1-\eta(h))], \eta''(h)=\E[X_1^2 \exp (hX_1-\eta(h))]-\E[X_1\exp(hX_1-\eta(h))]^2$.
And $\zeta $ is the solution to the implicit equation $\eta'(\zeta ) =\mu_2$ if $\mu_1<\mu_2$, then $\Lambda^*(\mu_2)=\zeta \mu_2 -\eta(\zeta)$.
We follow \cite{dembo1998large} and define a new probability measure $\widetilde P$ by $\frac{d\widetilde P}{dP}=\exp\left(\zeta x-\eta\left(\zeta\right)\right)$ and let $Z_i=\frac{X_i-\mu_2}{\sqrt{\eta''\left(\zeta\right)}}$. Then it follows that $Z_i$ are i.i.d. random variables satisfying $\E_{\widetilde P}[Z_i]=0, \E_{\widetilde P}[Z_i^2]=1, \E_{\widetilde P}[Z_i^3]\coloneqq z_3<\infty$.
We also define $\psi_m \coloneqq \zeta \sqrt{m \eta''\left(\zeta\right)}, W_m = \frac{\sum_{i=1}^m Z_i}{\sqrt{m}}$, $F_m$ is the cumulative distribution function of $W_n$ under measure $\widetilde P$, then we know that $\bar X_m=\mu_2 +\sqrt{\eta''\left(\zeta\right)/m}W_m$. Therefore, we can calculate
\begin{align}
    &\E_P [(\mu_2-\bar X_m) \1_{\{\bar X_m >\mu_2\}}]\\
    &=\E_{\widetilde P} [\exp(-m\left(\zeta \bar X_m-\eta(\zeta)\right)\cdot (-\sqrt{\eta''\left(\zeta\right)/m}W_m)\cdot \1_{\{\bar X_m >\mu_2\}}]\\
    &=-\exp\left(m(\zeta \mu_2-\eta(\zeta)) \right) \cdot \sqrt{\eta''\left(\zeta\right)/m} \cdot \E_{\widetilde P}[\exp\left( -m \zeta \sqrt{\eta''\left(\zeta\right)/m} W_m\right) \cdot W_m \1_{\{W_n >0\}}]\\
    &=-\exp \left(-m \Lambda^*(\mu_2) \right) \cdot \sqrt{\eta''\left(\zeta\right)/m} \cdot \int_0^{\infty} \exp(-\psi_m x)x dF_m.
\end{align}
Hence,
\begin{align}
    &J_m(\mu_2)\E_P [(\mu_2-\bar X_m) \1_{\{\bar X_m >\mu_2\}}]\\
    &= \zeta^2 \sqrt{2\pi m \eta''\left(\zeta\right)}m  \cdot \sqrt{\eta''\left(\zeta\right)/m} \cdot \int_0^{\infty} \exp(-\psi_m x)x dF_m\\
    &=\sqrt{2\pi} \psi_m^2 \int_0^{\infty} \exp(-\psi_m x)x dF_m(x)\\
    &=-\sqrt{2\pi} \psi_m^2 \int_0^{\infty} F_m(x) \exp(-\psi_m x)(-\psi_m x+1) dx.
\end{align}
Let $t=\psi_m x$, then
\begin{align}
    &J_m(\mu_2)\E_P [(\mu_2-\bar X_m) \1_{\{\bar X_m >\mu_2\}}]\\
    &=-\sqrt{2\pi} \psi_m \int_0^{\infty} F_m(t/\psi_m) \exp(-t)(-t+1) dt\\
    &=-\sqrt{2\pi} \psi_m \left[\int_0^{\infty} \left(F_m(t/\psi_m)-F_m(0) \right) \exp(-t)(-t+1) dx+\int_0^{\infty} F_m(0) \exp(-t)(-t+1) dt\right].
\end{align}
Note that $\int_0^{\infty}\exp(-t)(-t+1) dx=0$, therefore, we have
\begin{align}
    J_m(\mu_2)\E_P [(\mu_2-\bar X_m) \1_{\{\bar X_m >\mu_2\}}]=-\sqrt{2\pi} \int_0^{\infty} \psi_m \left(F_m(t/\psi_m)-F_m(0) \right) \exp(-t)(-t+1) dt.
\end{align}
We now discuss the lattice and non-lattice cases separately.

(a) When $X_1$ has a non-lattice law, the Berry-Esseen expansion \cite{feller2008introduction} of $F_m(x)$ yields
\begin{align}
    \lim _{m \to \infty}\left[\sqrt{m} \sup _{x}\left|F_{m}(x)-\Phi(x)-\frac{z_{3}}{6 \sqrt{m}}\left(1-x^{2}\right) \phi(x)\right|\right]=0,
\end{align}
where $z_3=\E_{\widetilde P}[Z_i^3]<\infty$.

Let $C_m \coloneqq -\sqrt{2\pi} \int_0^{\infty} \psi_m e^{-t} (-t+1)\left[ \Phi\left( \frac{t}{\psi_m}\right)+\frac{z_3}{6\sqrt{m}} \left(1-\left(\frac{t}{\psi_m}\right)^2 \right)\phi\left(\frac{t}{\psi_m}\right)-\Phi(0)- \frac{z_3}{6\sqrt{m}} \phi(0)\right] dt$. Note that $\psi_m=O(\sqrt{m})$, then we can get
\begin{align}
    \lim_{m \to \infty} \left|J_m(\mu_2)\E_{\mu} [(\mu_2-\bar X_m) \1_{\{\bar X_m >\mu_2\}}]-C_m \right|=0.
\end{align}
Moreover,
\begin{align}
    \lim_{m \to \infty} C_m &= \lim_{m \to \infty}(-\sqrt{2\pi}) \int_0^{\infty} \psi_m e^{-t} (-t+1)\left[ \Phi\left( \frac{t}{\psi_m}\right)-
    \Phi(0) \right] dt\\
    &\quad+\lim_{m \to \infty}(-\sqrt{2\pi}) \int_0^{\infty} \psi_m e^{-t} (-t+1) \left[\frac{z_3}{6\sqrt{m}} \left(1-\left(\frac{t}{\psi_m}\right)^2 \right)\phi\left(\frac{t}{\psi_m}\right)-\frac{z_3}{6\sqrt{m}} \phi(0) \right] dt\\
    &=\lim_{m \to \infty}(-\sqrt{2\pi}) \int_0^{\infty} \psi_m e^{-t} (-t+1)\left[ \Phi\left( \frac{t}{\psi_m}\right)-
    \Phi(0) \right] dt
\end{align}

Taylor expansion of $\Phi( \frac{t}{\psi_m})$ shows that $\Phi( \frac{t}{\psi_m})=\Phi(0)+\frac{t}{\psi_m}\phi(0)+O(\frac{t^2}{m})$, where $\phi$ is the density function of $N(0,1)$. Therefore by dominated convergence theorem we have
\begin{align}
    \lim_{m \to \infty} C_m = (-\sqrt{2\pi}) \int_0^{\infty}  e^{-t} (-t+1)t\phi(0) dt=\sqrt{2\pi}\phi(0)=1.
\end{align}
Thus,  we obtain
\begin{align}
    \lim_{m \to \infty} J_m(\mu_2)\E_P [(\mu_2-\bar X_m) \1_{\{\bar X_m >\mu_2\}}]=1.
\end{align}

(b) When $X_1$ has a lattice law, we have $Z_i=\frac{X_i-\mu_2}{\eta''\left(\zeta\right)} \in \{\frac{md}{\eta''\left(\zeta\right)}: m \in \mathcal{Z}\}$. From the Berry-Esseen expansion \cite{feller2008introduction}, we have
\begin{align}
\lim_{m \to \infty}\left[\sqrt{m} \sup_{x} \left| F_{m}(x)-\Phi(x)-\frac{z_{3}}{6 \sqrt{m}}\left(1-x^{2}\right) \phi(x)-\phi(x) g\left(x, \frac{d}{\sqrt{\eta''\left(\zeta\right) m}}\right) \right|\right]=0,
\end{align}
where $g(x,h)=\frac{h}{2}-(x \:\text{mod}\: h)$ if $(x \:\text{mod}\: h) \ne 0$ and $g(x,h)=-\frac{h}{2}$ if $(x \:\text{mod}\: h)=0$. Then,
\begin{align}
    &\lim_{m \to \infty} J_m(\mu_2)\E_P [(\mu_2-\bar X_m) \1_{\{\bar X_m >\mu_2\}}]\\
    &=1+\lim_{m \to \infty}(-\sqrt{2\pi}) \int_0^{\infty} \psi_m e^{-t} (-t+1)\left[\phi\left(\frac{t}{\psi_m}\right)g\left(\frac{t}{\psi_m}, \frac{\zeta d}{\psi_m}\right)-\phi(0)g\left(0, \frac{\zeta d}{\psi_m}\right) \right]dt.
\end{align}
Since $\psi_m g\left(\frac{t}{\psi_m}, \frac{\zeta d}{\psi_m}\right)=g(t,\zeta d)$, then
\begin{align}
    &\lim_{m \to \infty} J_m(\mu_2)\E_P [(\mu_2-\bar X_m) \1_{\{\bar X_m >\mu_2\}}]\\
    &=1+\lim_{m \to \infty}(-\sqrt{2\pi}) \int_0^{\infty} e^{-t} (-t+1)\left[\phi\left(\frac{t}{\psi_m}\right)g(t,\zeta d)-\phi(0)g(0, \zeta d) \right]dt\\
    &=1-\sqrt{2\pi} \phi(0) \int_0^{\infty} e^{-t} (-t+1)\left[g(t,\zeta d)-g(0, \zeta d) \right]dt\\
    &=1-\left(\sum_{n=0}^{\infty}e^{-n\zeta d}\right)\int_0^{\zeta d} e^{-t} (-t+1) (\zeta d-t)dt\\
    &=1-\left[\frac{1}{1-e^{-\zeta d}}\cdot \left(1-e^{-\zeta d}-\zeta d e^{-\zeta d} \right) \right]\\
    &=\frac{\zeta d e^{-\zeta d}}{1-e^{-\zeta d}}.
\end{align}
The proof is hence complete. 
\end{proof}

\section{Proof of Theorem \ref{thm:decayrate_subGaussian}} \label{sec:proof-thm2}
\begin{proof}
First, in the proof of Theorem~\ref{thm:decayrate_normal} (see \eqref{Pconverge:diff}), we have established that $|\hat{\mu}_1 - \mu_1| + | \hat{\sigma}_1^2 -\sigma_1^2|\stackrel{P}{\longrightarrow} 0$ as $m \rightarrow \infty$, where $\sigma_1^2= Var(X_1)$, the variance of the reward from Arm 1. This result does not depend on the assumption of Gaussian rewards. It then follows from the  continuous mapping theorem (Theorem 7.10 \cite{dasgupta2011probability}) that
\begin{align*}
  \hat \Lambda^*(\mu_2) =  \frac{1}{2\hat{\sigma}_1^2} (\hat{\mu}_1 - {\mu}_2)^2 \stackrel{P}{\longrightarrow} \frac{(\mu_2-\mu_1)^2}{2\sigma_1^2}, \quad \text{as $m \rightarrow \infty$}.
\end{align*}

Next, we show that $\Lambda^*(\mu_2) \geq \frac{(\mu_2-\mu_1)^2}{2a^2}$, where $\Lambda^*(\mu_2)$ is given in \eqref{eq:decay-rate-general}. Recall that $X_1 - \mu_1$ is sub-Gaussian with variance proxy parameter $a^2$, i.e. $\E\left[e^{\lambda (X_1-\mu_1)}\right] \leq \exp(\frac{a^2}{2} \lambda^2)$ for all $\lambda \in \mathbb{R}.$ It follows that when $\mu_1 <\mu_2$ we have the Chernoff-Hoeffding bound:
\begin{align}
    \mathbb{P}(\bar X_m \geq \mu_2) \leq e^{-\frac{m(\mu_2-\mu_1)^2}{2a^2}},
\end{align}
where $\bar X_m = \frac{1}{m} \sum_{i=1}^m X_i$ is the average of $m$ i.i.d sub-Gaussian rewards from Arm 1. Hence,
\begin{align*}
 \lim_{m \to \infty} \frac{1}{m}\log P(\bar X_m \ge \mu_2) \leq -\frac{(\mu_2-\mu_1)^2}{2 a^2}.
\end{align*}
On the other hand, from formula \eqref{BR:tailpro}, we know that
\begin{align}
    \lim_{m \to \infty} \frac{1}{m}\log P(\bar X_m \ge \mu_2)= \lim_{m \to \infty} \frac{1}{m} \log\left( \frac{c_*+o(1)}{\sqrt{2 \pi m \eta''(\zeta )}}\right)-\Lambda^*(\mu_2)=-\Lambda^*(\mu_2).
\end{align}
Therefore, we have
\begin{align*}
   \Lambda^*(\mu_2) \geq \frac{(\mu_2-\mu_1)^2}{2a^2} = \frac{(\mu_2-\mu_1)^2}{2 \sigma_1^2} \cdot \frac{ \sigma_1^2}{a^2}.
\end{align*}
When $\mu_1 > \mu_2,$ the proof is similar by studying $P(\bar X_m \le \mu_2),$ and hence we omit the details. The proof is complete.
\end{proof}

\end{document}